\documentclass[journal]{IEEEtran}

\usepackage{amssymb}
\usepackage{amsmath,epsfig}
\usepackage{booktabs} 
\usepackage{tabularx} 
\usepackage[colorlinks=true,linkcolor=black,urlcolor=blue,citecolor=black]{hyperref} 
\usepackage{lettrine} 
\usepackage{graphicx} 
\usepackage{subcaption} 
\usepackage[numbers]{natbib} 
\usepackage{amsmath} 
\usepackage{amssymb} 
\usepackage{lipsum} 
\usepackage[table]{xcolor} 
\usepackage{multirow} 
\usepackage{array} 
\usepackage{tabularx}
\usepackage{booktabs}
\usepackage{arydshln}  
\usepackage{orcidlink} 
\usepackage{hyperref}  
\usepackage{soul} 

\hypersetup{
    citecolor=blue
}

\newcounter{subsubsubsection}[subsubsection]
\renewcommand\thesubsubsubsection{\thesubsubsection.\arabic{subsubsubsection}}
\newcommand\subsubsubsection[1]{%
  \par\refstepcounter{subsubsubsection}%
  \paragraph{\thesubsubsubsection\hspace{1em}#1}}

\renewcommand{\baselinestretch}{1.0} 
\renewcommand{\thetable}{\Roman{table}} 

\begin{document}

\title{Semantic Segmentation in Satellite Hyperspectral Imagery by Deep Learning}

\author{
    Jon~Alvarez~Justo\orcidlink{0000-0003-4867-2693}\thanks{This work was supported by Norway Grants 2014 - 2021, under Project ELO-Hyp contract no. 24/2020, and by the Research Council of Norway grant no. 328724 (Green Platform) and no. 325961 (HYPSCI) (Corresponding author: Jon Alvarez Justo).\\Jon Alvarez Justo, Joseph L. Garrett, and Tor Arne Johansen are with the Department of Engineering Cybernetics, Norwegian University of Science and Technology, Trondheim, Norway (e-mail: jonalvjusto@gmail.com / jon.a.justo@ntnu.no; joseph.garrett@ntnu.no;   tor.arne.johansen@ntnu.no). Alexandru Ghi\c{t}\u{a}, Mariana-Iuliana Georgescu, and Radu Tudor Ionescu are with the Department of Computer Science, University of Bucharest, București, Romania (e-mail: alexandru.ghita94@gmail.com; georgescu\_lily@yahoo.com; raducu.ionescu@gmail.com).
    Daniel~Ková\v{c} is with the Department of Telecommunications, Brno University of Technology, Brno, Czech Republic (e-mail: xkovac41@vut.cz). Jesus Gonzalez-Llorente is with the Department of Aerospace Engineering, Nihon Universiy, Chiba, Japan (e-mail: jdgonzalezl@ieee.org).},
    Alexandru~Ghi\c{t}\u{a}\orcidlink{0009-0004-7286-4091}, 
    Daniel~Ková\v{c}\orcidlink{0000-0003-2701-1802}, 
    Joseph~L.~Garrett\orcidlink{0000-0001-8265-0661},~\IEEEmembership{Member,~IEEE,}
    Mariana-Iuliana~Georgescu\orcidlink{0000-0001-8277-1340},
    Jesus~Gonzalez-Llorente\orcidlink{0000-0001-6525-7657},~\IEEEmembership{Senior Member,~IEEE,}
    Radu~Tudor~Ionescu\orcidlink{0000-0002-9301-1950},~\IEEEmembership{Member,~IEEE,}
    and Tor~Arne~Johansen\orcidlink{0000-0001-9440-5989},~\IEEEmembership{Senior Member,~IEEE} 
}

\markboth{} {Justo \MakeLowercase{\textit{et al.}}: Semantic Segmentation in Satellite Hyperspectral Imagery by Deep Learning}

\maketitle


\begin{abstract}
Satellites are increasingly adopting on-board AI to optimize operations and increase autonomy through in-orbit inference. The use of Deep Learning (DL) models for segmentation in hyperspectral imagery offers advantages for remote sensing applications. In this work, we train and test 20 models for multi-class segmentation in hyperspectral imagery, selected for their potential in future space deployment. These models include 1D and 2D Convolutional Neural Networks (CNNs) and the latest vision transformers (ViTs). We propose a lightweight 1D-CNN model, 1D-Justo-LiuNet, which outperforms state-of-the-art models in the hypespectral domain. 1D-Justo-LiuNet exceeds the performance of 2D-CNN UNets and outperforms Apple's lightweight vision transformers designed for mobile inference. 1D-Justo-LiuNet achieves the highest accuracy (0.93) with the smallest model size (4,563 parameters) among all tested models, while maintaining fast inference. Unlike 2D-CNNs and ViTs, which encode both spectral and spatial information, 1D-Justo-LiuNet focuses solely on the rich spectral features in hyperspectral data, benefitting from the high-dimensional feature space. Our findings are validated across various satellite datasets, with the HYPSO-1 mission serving as the primary case study for sea, land, and cloud segmentation. We further confirm our conclusions through generalization tests on other hyperspectral missions, such as NASA's EO-1. Based on its superior performance and compact size, we conclude that 1D-Justo-LiuNet is highly suitable for in-orbit deployment, providing an effective solution for optimizing and automating satellite operations at edge.
\end{abstract}

\begin{IEEEkeywords}
Remote Sensing, Satellite Hyperspectral Imagery, Segmentation, Deep Learning, 1D-CNNs, 2D-CNNs, ViTs.
\end{IEEEkeywords}


\IEEEpeerreviewmaketitle


\section{Introduction}
\IEEEPARstart{A}{rtificial Intelligence (AI)} has seen a growing utilization aboard satellite platforms for remote sensing, benefiting Earth observation missions equipped with various technologies, from conventional RGB to Multispectral Imaging (MSI) and Hyperspectral Imaging (HSI). The importance of monitoring the Earth's environment with HSI technology is emphasized by the use of hyperspectral (HS) imagers in numerous contemporary satellite missions. Examples include $\Phi$-sat missions from the European Space Agency (ESA) \cite{esposito2019orbit}, \textit{PRecursore IperSpettrale della Missione Applicativa} (PRISMA) from the Italian Space Agency (ASI) \cite{cogliati2021prisma, cosmo2014future}, the \textit{Environmental Mapping and Analysis Program} (EnMAP) from the German Aerospace Center (DLR) \cite{guanter2015enmap}, and the \textit{HYPerspectral small Satellite for Ocean observation} (HYPSO) missions from the Norwegian University of Science and Technology (NTNU) \cite{grotte2021ocean}, with HYPSO-1 launched in 2022 and HYPSO-2 in August 2024. The importance of HS imagers on satellites is further highlighted by upcoming missions such as the \textit{The Copernicus Hyperspectral Imaging Mission for the Environment} (CHIME) from ESA \cite{nieke2018towards, nieke2023copernicus}, the \textit{Plankton, Aerosol, Cloud, ocean Ecosystem} (PACE), the \textit{Geosynchronous Littoral Imaging and Monitoring Radiometer} (GLIMR), the \textit{Surface Biology and Geology} (SBG) from the National Aeronautics and Space Administration (NASA) \cite{Die23}, ASI/PRISMA Second Gen \cite{ansalone2023prisma}, as well as NTNU/HYPSO-3 \cite{langer2023robust}.
HS missions offer great potential due to their detailed spectral characterization of Earth's features, but they also share a common limitation with other optical remote sensing technologies: the challenge of cloud formation in the atmosphere obstructing the satellites' capacity to monitor targets of interest on the Earth's surface. In this context, techniques such as semantic segmentation proves valuable in a broad range of space applications, e.g., from cloud recognition to wildfire detection \cite{giuffrida2020cloudscout, giuffrida2021varphi, salazar2022cloud, pitonak2022cloudsatnet, spiller2022wildfire, lebedeff2020board, gjendem2023real, roysland2023hyperspectral}. Segmentation can indeed serve as a cloud detection technique, categorizing image pixels as either cloudy or non-cloudy, essentially performing binary classification. As a mission example, after the launch in 2020 of the $\Phi$-sat-1 HS mission part of ESA's Earth Observation Programmes \cite{esposito2019orbit}, a significant milestone was marked in 2021 when, for the first time, on-board AI was introduced in a satellite for object detection, with the initial focus being the detection of clouds aboard the satellite. $\Phi$-sat-1 marked the first in-orbit inference at edge with its convolutional CloudScout Deep Learning (DL) model for cloud detection \cite{giuffrida2020cloudscout, pitonak2022cloudsatnet}. The mission's imager, HyperScout-2, captures the Visible and Near-Infrared (VNIR) and Thermal Infrared (TIR) spectral ranges, and subsequently deploys data processing via AI models in its on-board configurable platform facilitating updates for its algorithms in flight \cite{esposito2019orbit, giuffrida2021varphi}. These in-orbit updates rely on an embedded configurable Commercial Off-The-Shelf (COTS) platform, which is based on a Vision Processing Unit (VPU) \cite{giuffrida2021varphi, kothari2020final}. Furthermore, beyond the $\Phi$-sat-1 mission, the literature \cite{salazar2022cloud} also extends cloud segmentation to image data of the missions Sentinel-2B (MSI) and FACSAT-1 (RGB) \cite{Giovanni2019}. For instance, the state of the art \cite{salazar2022cloud} deploys a satellite data priority system to compute the cloud coverage level via DL-based segmentation on microcontrollers (MCUs), which determines the image downlink priority. Furthermore, the literature demonstrates that cloud detection can enhance data management aboard satellite platforms as presented next. First, selective HSI compression methods may be applied to cloud segments via CCSDS-123.0-B-2 to reduce their encoding rate \cite{hernandez2021ccsds, birkeland2022board}, while other methods such as Compressed Sensing (CS) can also be employed to handle cloudy data, where CS proves particularly advantageous for satellites, as it transfers computational complexity from orbit to ground-based reconstruction \cite{justo2022comparative, justo2022study}. Therefore, segmentation can lead to selective compression reducing the bit rate of data segregated as cloudy, or its absolute removal by data screening \cite{giuffrida2021varphi}. This enables the satellite to downlink, for instance, only valuable non-cloudy data, thus optimizing the usage of the often limited communication channel \cite{salazar2022cloud}. On-board cloud segmentation can also play a role in reducing decision-making latency, increasing satellite autonomy with more optimal operations \cite{giuffrida2021varphi, salazar2022cloud, spiller2022wildfire, furano2020towards, wang2019satellite, raoofy2022benchmarking, ghiglione2021machine}. Furthermore, from a broader perspective, the models first trained for cloud binary segmentation can be retrained to solve more complex problems instead, e.g., multi-class segmentation, where each image pixel is assigned to one of various categories. To further illustrate the importance of AI-driven cloud and multi-class segmentation for space applications, we next employ a particular small satellite mission as a case study. In this work, we employ the Earth observation CubeSat HYPSO-1 \cite{grotte2021ocean}. In this mission, reducing decision-making latency is especially important for time-sensitive remote sensing applications, such as early event warnings. HYPSO-1 focuses on timely detection of marine phenomena such as Harmful Algal Blooms (HABs), which can lead to significant mortality of farmed salmon. In-orbit segmentation for adequate on-board data management can play a crucial role enabling prompt management responses to mitigate adverse effects in such scenarios. In the context of multi-class segmentation, by segregating the data along various categories of interest, on-board AI can activate other subsequent agents in observational hierarchies monitoring the environment at multiple spatial, spectral and temporal scales with e.g. drones and marine vessels \cite{grotte2021ocean, Bakken2023, dallolio2021satellite}, where the HYPSO-1 satellite may serve as the space segment of an observational pyramid. However, handling increasingly advanced tasks on satellites, such as segmentation of growing complexity, is feasible only when the COTS components within the satellite's payload are appropriately designed and optimized. Some COTS platforms provide benefits by accelerating, with high power efficiency, the repetitive operations commonly found in DL networks \cite{esposito2019orbit, pitonak2022cloudsatnet, kothari2020final, furano2020towards}. Indeed, to enable AI applications in satellites, the literature explores various COTS hardware, from the conventional MCUs \cite{salazar2022cloud} to the high-performance VPUs aboard $\Phi$-sat-1, Field-Programmable Gate Arrays (FPGAs) on board HYPSO missions, Tensor Processing Units (TPUs), Graphics Processing Units (GPUs), amog other dedicated accelerators \cite{esposito2019orbit, langer2023robust, giuffrida2020cloudscout, furano2018roadmap, antonini2019resource, li2020survey}.

Despite the extensive use of 2D Convolutional Neural Networks (CNNs) in state-of-the-art semantic segmentation for hyperspectral imaging, our primary research question is: How can we compress deep models to create lighter versions for future in-orbit deployment, while not only maintaining but also improving performance compared with state-of-the-art methods, from standard UNet architectures to the latest advanced vision transformers? Furthermore, to the best of our knowledge, the existing literature in the field of cloud detection on board satellite platforms does not compare how different shallow and deep learning models, including distinctions such as 1D vs.~2D-CNN-based models and Vision Transformers (ViT), influence the performance and complexity of cloud segmentation. Additionally, the field also overlooks the effects of reducing the number of spectral channels, particularly in MSI/HSI data, on models for cloud segmentation. The existing literature lacks comparisons of key aspects, such as whether on-board AI for cloud detection, is better applied to raw data or calibrated data. It does not discuss how L1b calibration impacts segmentation results and does not provide justification for the optimal stage in the on-board data pipeline for image segmentation at edge. Moreover, within the literature pertaining to cloud detection in HS satellite imagery, the primary emphasis naturally lies on binary segmentation, distinguishing between the presence of clouds and non-cloudy data; in fact, as earlier mentioned, there are ranking systems that assess cloud levels based on the segmented outcomes. Nevertheless, none of the systems at the time of writing perform multi-class semantic segmentation for the simultaneous detection of clouds, sea, and land categories, which typically account for most of the variance in HS satellite data. The absence of sea-land-cloud segmentation systems impedes, as a result, the development of applications such as priority ranking systems that can integrate the three categories into a unified ranker. In this work, we tackle the mentioned research gaps and utilize the HYPSO-1 mission as a case study, using its ``Sea-Land-Cloud-Labeled Dataset'' openly accessible in our prior article \cite{justo2023open} (see data availability section at the end of this paper).

Following a methodology inspired by the data science CRISP-DM approach (CRoss Industry Standard Process for Data Mining) to ensure robust and reliable results, we train and compare 19 different machine learning (ML) models. For further comparison, we will train an additional model, bringing the total to 20 models in this study. This extends the segmentation model baseline for HYPSO-1 first established in \cite{justo2023open}, and our research findings make contributions to the field as follows:

\begin{itemize}
    \item We propose and fine-tune four new 1D-CNN and 2D-CNN models, with the aim of adapting them for on-board deployment on satellites. This optimization entails reducing the number of model parameters to achieve highly compact lightweight models, while maintaining performance.
    \item In the context of multi-class HS segmentation, we demonstrate that 1D-CNN models exhibit superior performance when compared with 2D-CNN UNets and FastViT. Indeed, our lightweight 1D-CNN models, 1D-Justo-LiuNet and 1D-Justo-HuNet, consistently outperform the state of the art. Our results are confirmed by generalization tests on different satellites.

    \item We substantiate that segmenting L1b radiance data, as opposed to using raw sensor data, slightly enhances the segmentation of top-performing models.
    \item We explore the impact of channel reduction on sea-land-cloud segmentation models.
    \item Finally, we propose a case example consisting of an image ranking system intended for downlink prioritization based on sea, land, and cloud coverage.
\end{itemize}

\section{Methodological Background}
\label{Section:methodology}

\subsection{Datasets: Labeled Satellite Imagery}

We perform experiments on three datasets. The first one is the Sea-Land-Cloud-Labeled Dataset from HYPSO-1 \cite{justo2023open}\footnote{\url{https://ntnu-smallsat-lab.github.io/hypso1_sea_land_clouds_dataset/}}. The dataset consists of pixel-level labels for 25 million pixels across 38 scenes captured by HYPSO-1, where each HS image has a spatial dimension of 956 $\times$ 684 pixels. Each pixel contains a spectral signature of 120 channels covering the 400 to 800 nm range in the visible and VNIR spectrum, with an approximate spectral resolution of 5 nm and a spatial resolution of approximately 100 m $\times$ 600 m in the across- and along-track directions \cite{grotte2021ocean, Bakken2023}. The dataset consists of diverse scenes from various environments and provides raw data and L1b calibrated radiance for comparison with other satellite missions. 
The dataset includes expert pixel-level annotations for three classes: \textit{sea}, \textit{land}, and \textit{clouds}. 
The data exhibits a slight class imbalance with 37.01\% pixels representing sea, 40.14\% representing land, and 22.85\% representing clouds/overexposed areas.

Our method involves training 19 distinct models that are considered of interest for in-orbit deployment, with a particular emphasis on DL models, as they are often better suited to capture more complex data relationships. To train the models, we maintain the same data split used in our previous work~\cite{justo2023open}, allocating 30 HS images for training (79\%), 3 images for validation (8\%), and 5 images for testing (13\%). Furthermore, we employ a deployment set of 30 additional HS images, from the same sea-land-cloud dataset, to test the models ability to generalize to further new unseen data. In summary, we test on 35 different scenes in total from the deployment and test sets. The scenes comprise coastal zones, forests, dry regions, cloudy areas, and more. The deployment set is provided in the supplementary material of this article (see data availability section), where additional details, such as the specific image identifiers for the data split, are also included.

Furthermore, to assess the generalization of the models in other satellites, we test them on two additional datasets. First, we perform experiments using 35 calibrated HS images from NASA's Earth Observing-1 (EO-1) mission~\cite{ungar2003overview}\footnote{\url{https://earthexplorer.usgs.gov/}}. Second, we conduct experiments on Satellite Imagery of Dubai (SID) dataset\footnote{\url{https://humansintheloop.org/resources/datasets/semantic-segmentation-dataset-2/}}. For EO-1, the mission utilizes the Hyperion sensor~\cite{numata2011analyzing}, delivering high spatial (30 m) and spectral (10 nm) resolutions. Our EO-1 dataset includes expert ground-truth annotations with labels: \textit{no cloud}, \textit{thin cloud}, and \textit{thick cloud}. Further details on this dataset can be found in our article~\cite{kovac2024deep}. In the case of the SID dataset, it comprises 72 RGB aerial images of Dubai gathered by the Mohammed Bin Rashid Space Center (MBRSC). The images come with annotations at the pixel level with 6 class labels: \textit{buildings}, \textit{roads}, \textit{land}, \textit{vegetation}, \textit{water}, and \textit{other}.

\subsection{Channel Reduction Preprocessing}
\label{Section:Methodology_for_Dimensionality_Reduction}

To begin, we emphasize that the main goal of our work is not to find the optimal dimensionality reduction for our application. Instead, we aim to identify any features in the data that could impact our models and, more specifically, to examine how segmentation results are influenced when dimensionality is significantly reduced.

For data preprocessing, we first identify in the HYPSO-1 dataset unusual channel features that may require reduction. In the sea-land-cloud dataset \cite{justo2023open}, we observe that the first four channels with wavelengths $\lambda$=388, 391, 395, and 399 nm (visible blue spectrum) exhibit data samples equal to 0 for all the dataset images. We observe that this phenomenon occurs only for the L1b radiance, however, it is not present in the raw sensor data. This suggests that the occurrence of zero values in the first channels is a result of the radiometric calibration procedures applied by \cite{justo2023open}, following \cite{henriksen2022pre}. Furthermore, we observe that the dataset, near its wavelength $\lambda$=760 nm (oxygen A-band within the NIR spectrum), presents a sharp decline in light intensity which is attributed to the significant light absorption by the atmosphere's oxygen at this wavelength \cite{wark1965absorption, bouter2007satellite}. In summary, we exclude first wavelengths $\lambda$=388, 391, 395, and 399 nm, and we also exclude the wavelengths in close proximity to $\lambda$=760 nm, namely, $\lambda$=758, 761, 765, and 768 nm. As a result, the dataset's original 120 channels are reduced to 112. The 112 channels, after undergoing min-max normalization, are the ones we employ for training the ML and DL models in the subsequent methodology. Furthermore, this work also examines the impact of significantly reducing the number of channels on sea-land-cloud HSI segmentation. Therefore, we use the 112-channel data to create a new dataset containing only 3 channels, since the legacy algorithms for cloud detection in HS satellite imagery often select 3 channels to mimic RGB image processing \cite{giuffrida2020cloudscout, giuffrida2021varphi, salazar2022cloud, pitonak2022cloudsatnet}. Details about the dimensionality reduction algorithm used, which is beyond the scope of this work, can be found in Appendix \ref{Appendix:PCA_redution}. In the appendix, we explain that the 3 channels selected correspond to 412.72 nm for the blue spectrum, 699.61 nm for the green and red spectra, and 747.77 nm for the NIR spectrum.

\subsection{Semantic Segmentation Modeling}

\subsubsection{Processing of 1D vs.~3D Data}

The models presented in this work handle the data in either one-dimensional (1D) or three-dimensional (3D) formats. On the one hand, when we refer to 1D processing, it implies that the models make predictions solely based on spectral characteristics, disregarding any nearby spatial context. Examples of such models include 1D-CNNs based on 1D convolutions, where the input consists of a spectral signature with no spatial context. Henceforth, we refer to these types of models as 1D models. However, when we refer to 3D processing, it signifies that the models concurrently analyze multiple adjacent spatial pixels, incorporating the hyperspectral datacube with both spectral (1D) and spatial (2D) information to predict the class for each pixel. Models such as 2D-CNNs, that we refer to henceforth as 2D models serve as examples, where the input is a 3D image containing both spatial and spectral data. Convolutional layers in 2D-CNN models operate by first applying 2D convolutions individually between each spectral channel and its respective neural kernel component. Subsequently, they aggregate the 2D convolutional results stacked in the spectral dimension to produce a final 2D feature map. Finally, we describe in Appendix \ref{Appendix:data_dimensions_arrengement} the arrangement of the data dimensions required for training and inference. In the appendix, we provide details on how the images are divided into non-overlapping 3D patches to train the 2D-CNNs and reduce computational complexity by processing each pixel only once, while using a small patch size of 48 $\times$ 48. This size ensures integer spatial dimensions through pooling layers halving dimensions of feature maps, while being also especially convenient for reducing computational complexity in orbit compared to processing larger patches.

\subsubsection{Baseline Segmentation Models}

As a starting point, we begin using simple baseline models consisting of traditional ML algorithms, as they are often more interpretable than deep neural networks. For convenience, during this initial phase, the models process 1D data (spectral signatures), which is a simpler processing strategy compared to handling 3D image patches. The initial ML baseline consists of Stochastic Gradient Descent (referred henceforth as 1D-ML/SGD), Gaussian Na\"{i}ve Bayes (1D-ML/NB), Linear Discriminant Analysis (1D-ML/LDA), and Quadratic Discriminant Analysis (1D-ML/QDA). Next, we incorporate multiple state-of-the-art DL models for 3D patch processing, such as NU-Net \cite{park2020rgb} and NU-Net-mod \cite{salazar2022cloud}, previously employed in satellite imagery for cloud detection in RGB (FACSAT-1) and MSI (Sentinel-2B). Furthermore, we incorporate additional DL models, namely C-UNet and C-UNet++, distinguished by the authors \cite{bahl2019low} as lightweight for their minimal number of parameters and low computational requirements, making them well-suited for deployment on satellites. C-UNet and C-UNet++ are compact versions of the standard encoder-decoder U-Net architecture \cite{ronneberger2015u}. The subsequent literature \cite{tagestad2021hardware} building upon \cite{bahl2019low}, explores C-UNet and C-UNet++ models suggesting their possible use for semantic segmentation in HS missions such as HYPSO-1. However, the method in \cite{tagestad2021hardware} involves training on only two limited HS images, the conventional Pavia Centre and AeroRIT scene, which use sensors different from HYPSO-1 and do not include any cloud or oceanic data, which is of particular interest for ocean observation in HYPSO-1. In our work, we refer to the compact models in \cite{tagestad2021hardware} as 2D-CUNet and 2D-CUNet++, and to their further reduced versions in \cite{tagestad2021hardware} as 2D-CUNet Reduced and 2D-CUNet++ Reduced. Moreover, we add models from \cite{netteland2022exploration}, which are referred to by Netteland \cite{netteland2022exploration} as the FAUBAI models, with reference to the FAUBAI project that focuses on segmentation for HS imagery for its possible future use aboard satellites. We refer to the networks in \cite{netteland2022exploration} as 2D-UNet FAUBAI and to its reduced version as 2D-UNet FAUBAI Reduced. Although Netteland \cite{netteland2022exploration} recommends any of these two networks for tasks such as cloud segmentation in HS satellite imagery, we note that similarly to \cite{tagestad2021hardware}, Netteland \cite{netteland2022exploration} conducted model testing only on a restricted set of four conventional HS images, namely, Pavia Centre, Pavia University, Salinas, and Indian Pines, which do not contain cloud or oceanic data.

Finally, we include one of the most recent and advanced lightweight models as a baseline. Fast Hybrid Vision Transformer (FastViT) is a state-of-the-art model introduced by Apple in 2023~\cite{vasu2023fastvit} for processing image patches in tasks such as semantic segmentation, object detection, and image classification. FastViT is designed as a lightweight model optimized for edge inference on mobile devices and has been tested on platforms such as the iPhone 12 Pro. It introduces a novel token-mixing operator that utilizes structural reparameterization to enhance performance by eliminating skip connections, reducing memory access and footprint to optimize efficiency for mobile environments. Vasu et al.~\cite{vasu2023fastvit} proposed 8 architectures of various depths. From the set of architectures, we choose the shallowest one FastViT-T8, the deepest one (FastViT-MA36), and a middle-sized one (FastViT-S12).

    \subsubsection{Customised Semantic Segmentation Models}
    
        \subsubsubsection{Review of 1D-CNNs: Liu, Hu, and Lucas CNNs}

The \textit{Liu} \cite{liu2018transfer}, \textit{Hu} \cite{hu2015deep} and \textit{Lucas} \cite{soil_texture_one_dimensional_CNN_2019} 1D-CNN models have been used before for data analysis in soil spectroscopy to segment various soil texture classes in HS data. Additionally, in our previous article \cite{justo2023open}, we verified that a modified version of \textit{LiuNet} achieved high performance in satellite HS imagery, surpassing the state-of-the-art in soil spectroscopy \cite{soil_texture_one_dimensional_CNN_2019}. In this work, we find \textit{LiuNet}, \textit{HuNet}, and \textit{LucasCNN} of interest for in-orbit inference due to their simple 1D processing approach. Therefore, assuming 112 spectral channels and 3 classes to segment, in Table \ref{Table:model_hyperparameters}, we show the parameter counts of these three models. The parameter counts are in the order of tens of thousands, with \textit{LiuNet} having the lowest number of parameters among the three. Following our pursuit of further reducing the parameter count to merely a few thousands, more convenient for inference aboard satellites, in the next sections we accordingly modify the soil models in \cite{soil_texture_one_dimensional_CNN_2019}, while aiming to preserve their performance.

\begin{table*}[t] 
\setlength\tabcolsep{2.0pt}
    \centering
    \caption{Overview of hyperparameters and parameter count for 1D-CNN segmentation models. }

    \scalebox{0.8}{
    \begin{tabular}{|l||c|c|c|}

        \hline  
        Model & Parameter Count & Layers & Parameters \\ \hline 
        
        \textit{LiuNet} & 22,755 & 
            \parbox[t]{6.3cm}{Hidden: 4 1D Convolutions \& 4 Max Poolings
                            \\ Output: Flatten \& 1 Dense} 
                            & 
                \parbox[t]{6cm}{
                            Hidden/Convolutions: \textit{N=32,32,64,64}; \textit{K=3}; \textit{Act.=ReLU} \\ 
                            Hidden/Poolings: \textit{÷2} \\
                            Output/Dense: \textit{N=3}; \textit{Act.=Softmax}
                }
        \\ \hdashline 

      \textbf{1D-Justo-LiuNet} & 4,563 & 
            \parbox[t]{6.3cm}{Hidden: 4 1D Convolutions \& 4 Max Poolings
                            \\ Output: Flatten \& 1 Dense} 
                            & 
                \parbox[t]{6cm}{
                            Hidden/Convolutions: \textit{\textbf{N=6,12,18,24}}; \textit{\textbf{K=6}}; \textit{Act.=ReLU} \\ 
                            Hidden/Poolings: \textit{÷2} \\
                            Output/Dense: \textit{N=3}; \textit{Act.=Softmax}
                }
        \\ \hline \hline

      \textit{HuNet} & 100,663 & 
            \parbox[t]{6.3cm}{Hidden: 1 1D Convolution \& 1 Max Pooling
                            \\ Output: Flatten \& 2 Dense} 
                            & 
                \parbox[t]{6cm}{
                            Hidden/Convolution: \textit{N=20}; \textit{K=12}; \textit{Act.=Tanh} \\ 
                            Hidden/Pooling: \textit{÷2} \\
                            Output/Dense 1: \textit{N=100}; \textit{Act.=Tanh} \\
                            Output/Dense 2: \textit{N=3}; \textit{Act.=Softmax}
                }
        \\ \hdashline

      \textbf{1D-Justo-HuNet} & 9,543 & 
            \parbox[t]{6.3cm}{Hidden: 1 1D Convolution \& 1 Max Pooling
                            \\ Output: Flatten \& 2 Dense} 
                            & 
                \parbox[t]{6cm}{
                            Hidden/Convolution: \textit{\textbf{N=6}}; \textit{\textbf{K=9}}; \textit{Act.=Tanh} \\ 
                            Hidden/Pooling: \textit{÷2} \\
                            Output/Dense 1: \textit{\textbf{N=30}}; \textit{\textbf{Act.=ReLU}} \\
                            Output/Dense 2: \textit{N=3}; \textit{Act.=Softmax}
                }
        \\ \hline  \hline

      \textit{LucasCNN} & 80,155 & 
            \parbox[t]{6.3cm}{Hidden: 4 1D Convolutions \& 4 Max Poolings
                            \\ Output: Flatten \& 3 Dense} 
                            & 
                \parbox[t]{6cm}{
                            Hidden/Convolutions: \textit{N=32,32,64,64}; \textit{K=3}; \textit{Act.=ReLU} \\ 
                            Hidden/Poolings: \textit{÷2} \\
                            Output/Dense 1: \textit{N=120}; \textit{Act.=ReLU} \\
                            Output/Dense 2: \textit{N=160}; \textit{Act.=ReLU} \\
                            Output/Dense 3: \textit{N=3}; \textit{Act.=Softmax}
                }
        \\ \hdashline

      \textbf{1D-Justo-LucasCNN} & 25,323 & 
            \parbox[t]{6.3cm}{Hidden: \textbf{1 1D Convolution} \& \textbf{1 Max Pooling}
                            \\ Output: Flatten \& 3 Dense} 
                            & 
                \parbox[t]{6cm}{
                            Hidden/Convolution: \textit{\textbf{N=16}}; \textit{\textbf{K=9}}; \textit{\textbf{Act.=Tanh}} \\ 
                            Hidden/Pooling: \textit{÷2} \\
                            Output/Dense 1: \textit{\textbf{N=30}}; \textit{\textbf{Act.=Tanh}} \\
                            Output/Dense 2: \textit{\textbf{N=5}}; \textit{\textbf{Act.=Tanh}} \\
                            Output/Dense 3: \textit{N=3}; \textit{Act.=Softmax}
                }
        \\ \hline \hline \hline

      \textbf{2D-Justo-UNet-Simple} & 7,641 & 
            \parbox[t]{6.3cm}{Hidden: \textbf{3 2D Convolutions} \& \textbf{2 Max Poolings} \& \textbf{2 Upsamplings}  \\
                        Output: 1 2D Convolution \\
                            * \textbf{Batch normalization after all 2D Convolutions}} 
                            & 
                \parbox[t]{6cm}{
                            Hidden/Convolutions: \textbf{\textit{N=6,12,6}}; \textit{K=3x3}; \textit{Act.=ReLU} \\ 
                            Hidden/Poolings: \textit{÷2} \\
                            Hidden/Upsamplings: \textit{x2} \\
                            Output/Convolution: \textit{N=3}; \textit{K=3x3}; \textit{Act.=Softmax}
                }
        \\ \hline

    \end{tabular}
    }
    \label{Table:model_hyperparameters} 
\end{table*}

        \subsubsubsection{Ablation Study}
 
To adapt the soil models for deployment in space by reducing their trainable parameters while not degrading their performance, we initially investigate the relevance of each neural layer and subsequently conduct hyperparameter fine-tuning. In the first column of Table \ref{Table:model_hyperparameters}, we list the state-of-the-art soil models in italic font, while our customized models are listed in bold. We also differentiate our models by the addition of the term ``Justo'' to their names. In the subsequent columns of the table, for our models, we utilize bold font to highlight the hyperparameters that we modify with respect to the soil models.

Taking into account the promising nature of 1D-CNNs as lightweight models for on-board deployment, we focus the hyperparameter optimization exclusively on them, and we omit the optimization for 2D-CNN models. The modifications to the hyperparameters in Table~\ref{Table:model_hyperparameters} further discussed in the following section, are a direct outcome of the hyperparameter fine-tuning process. Namely, we begin by assigning them intuitive values that can yield acceptable performance. Then, we conduct a search using a hyperparameter grid, based on the initial values, to identify the optimal combination of hyperparameters. 
To accommodate the high dimensionality of HSI data, we randomly choose only 2 million spectral signatures (approximately 10\% of the training set) to reduce the time needed for the grid search. Additionally, while evaluating the grid models through cross-validation, we restrict the process to only two fold permutations to further decrease computation time. 

        \subsubsubsection{Introducing New Lightweigth 1D-CNNs:\\ 1D-Justo-LiuNet, 1D-Justo-HuNet and 1D-Justo-LucasCNN}

The aforementioned hyperparameter optimization process aims to achieve optimal accuracy, but we emphasize that the minimization of parameter counts for optimal in-orbit deployment is still the primary focus of our method. In this context, in our prior article \cite{justo2023open}, we presented a modified version of the \textit{LiuNet} network resulting in 124,163 parameters, which may be considered large for in-flight deployment. However, our results demonstrated superior performance when compared with the original \textit{LiuNet} model \cite{soil_texture_one_dimensional_CNN_2019}. We consider that the large parameter count in \cite{justo2023open} likely contributed to the significant performance improvement over the state of the art. Nevertheless, in this article, our goal is to once again outperform the state of the art, while also reducing the parameter count, in contrast to our previous work \cite{justo2023open}. \begin{figure}[t]
    \centering 
\includegraphics[width=0.45\textwidth]{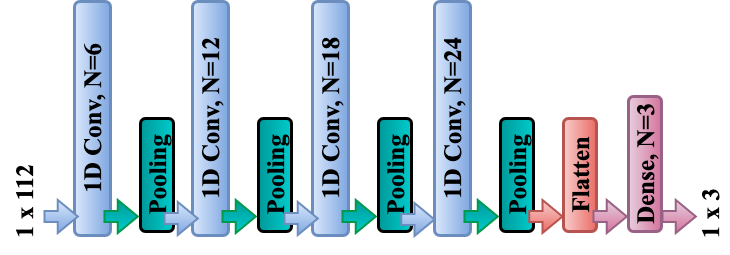} 
    \caption{Lightweight 1D-Justo-LiuNet architecture for on-board segmentation. The block diagram employs 112 input channels for a single pixel. The network generates three detection probabilities, corresponding to sea, land, and clouds, respectively.}
    \label{Fig:justoliunet_architecture} 
\end{figure}Indeed, we aim to achieve a parameter count even lower than the count of the \textit{LiuNet} model in Table \ref{Table:model_hyperparameters}. Therefore, we next present the 1D-Justo-LiuNet in Fig.~\ref{Fig:justoliunet_architecture}, with 80\% reduction in parameters compared with \textit{LiuNet} (22,755) \cite{soil_texture_one_dimensional_CNN_2019}. 1D-Justo-LiuNet is reduced to only 4,563 parameters, as shown in Table \ref{Table:model_hyperparameters}. This reduction is accomplished by first following the approach in our prior article \cite{justo2023open}, suggesting to start with a base seed of convolutional kernels and progressively increase the kernel counts (denoted as \textit{N} in Table \ref{Table:model_hyperparameters}) based on the seed multiplied by factors of 2, 3, and 4 as the network deepens. However, in this work, we start with a smaller seed of 6 kernels, instead of 32 as in \cite{justo2023open}, to achieve the substantial 80\% reduction in the model's parameter count. Moreover, as we did in our previous publication, we double the kernel size (denoted as \textit{K} in Table \ref{Table:model_hyperparameters}) compared with \textit{LiuNet} \cite{soil_texture_one_dimensional_CNN_2019} in order to achieve a larger receptive field for input samples and compensate for the reduced number of kernels, thereby ensuring that the performance is not adversely affected.

For the \textit{HuNet} model in \cite{soil_texture_one_dimensional_CNN_2019}, the high parameter count in Table \ref{Table:model_hyperparameters} is mainly a result of a fully connected layer with numerous neurons preceding the output classification layer. We refine the model, naming it 1D-Justo-HuNet, by reducing 70\% the neural units in the fully connected layer. By implementing additional modifications such as reducing the number of convolutional kernels and adjusting their kernel size, we attain a significant 90.50\% total reduction in parameter count (9,543). Regarding \textit{LucasCNN} \cite{soil_texture_one_dimensional_CNN_2019}, our model, 1D-Justo-LucasCNN, follows the same approach applied to \textit{HuNet} to reduce the parameter count, i.e., we significantly reduce the neural units in the two dense layers before the classification layer. Specifically, \textit{LucasCNN} has 280 neural dense units, while 1D-Justo-LucasCNN reduces the units by 87\% to just 35 neurons. Furthermore, 1D-Justo-LucasCNN includes other adjustments, such as transforming the 4-scale \textit{LucasCNN} into a single-scale network, significantly reducing the number of convolutional kernels while tripling the kernel size, and employing a hyperbolic neural activation instead of a linear function.  Table \ref{Table:model_hyperparameters} illustrates a 68.41\% reduction in parameter count (25,323) achieved by 1D-Justo-LucasCNN. The architectural details (as in Fig.~\ref{Fig:justoliunet_architecture}) of the 1D-Justo-HuNet and 1D-Justo-LucasCNN are available in our supplementary material, which also include the software codes to reproduce the models and an interactive Python notebook showing how model inference is performed.

        \subsubsubsection{Introducing a New Lightweight 2D-CNN:\\2D-Justo-UNet-Simple}

Finally, to achieve lightweight models with reduced parameters, not only for 1D-CNN models as in previous section, but also for 2D-CNN architectures, we introduce 2D-Justo-UNet-Simple, a model with 7,641 parameters, as shown in Table \ref{Table:model_hyperparameters}. The 2D-Justo-UNet-Simple consists of a compressed version of the encoder-decoder U-Net architecture with only 2 scales (as opposed to the standard 4-scale U-Net originally proposed for biomedical image segmentation in \cite{ronneberger2015u}), with a bottleneck level consisting of only 12 convolutional kernels. Additionally, following each 2D convolution operation and before pooling or upsampling, we insert batch normalization before the rectified linear activation to enhance model generalization and mitigate potential gradient vanishing or exploding effects during training. We note that in the decoder's expanding path, we employ upsampling layers as an alternative to utilizing operations such as transposed convolutions. While transposed convolutions may enhance performance, they also lead, however, to an increase in the number of trainable parameters and complexity. The software code and architectural details (as in Fig.~\ref{Fig:justoliunet_architecture}) of the 2D-Justo-UNet-Simple are also provided in the supplementary material of this article.

    \subsubsection{Training Procedure}
    \label{Section:methodology_training_procedure}

The DL models (implemented with Keras) are trained on an NVIDIA RTX A4000 GPU to increase the training speed, while the ML models (implemented with scikit-learn) are trained on a 13th Gen Intel(R) Core (TM) i9-13900K CPU, which results in a slower training process relying on CPU computation. We employ supervised learning to train the ML and DL models, with a special focus on the training process of DL models. This is crucial because, due to their larger number of parameters, DL models are more susceptible to overfitting than ML models.

We start by using the 112-channel dataset to train the 1D and 2D models. However, when working with the 3-channel dataset, we exclusively focus on training 2D models instead of 1D models. This decision is primarily based on the fact that 1D models do not take spatial information into account and rely solely on the spectral features, which, in this case, comprise only 3 channels. Furthermore, 1D-CNN models can face challenges related to the diminishing channel features as the network depth increases.

In our training procedure, we initially select high-level parameters such as the number of epochs and the batch size, followed by the choice of the lower-level hyperparameters previously discussed. In terms of high-level parameters, for both 1D-CNN and 2D-CNN models, we find that our dataset benefits from small batch sizes, increasing the frequency of updates in trainable parameters, but also extending the training time. As regards to batch size during inference, Keras dynamically selects the batch size for optimal memory efficiency. In addition, we find that it is advisable to utilize a small number of training epochs to prevent overfitting. Specifically, for 1D-CNNs, we use 2 epochs and a batch size of 32 pixels, while for 2D-CNNs, we opt for 3 epochs and a batch size of 4 patches. Furthermore, we employ the Adam optimizer with its default parameters to minimize the categorical cross-entropy loss, while using accuracy as evaluation metric in the training and validation sets. To evaluate the model's performance at the end of each training epoch, we focus on the overall validation accuracy (validation set imbalance: 51.25\%, 28.13\%, and 20.62\% for sea, land, and clouds, respectively), and we aim to achieve an empirical validation accuracy of around 0.90 during training. After completing the last training epoch, we evaluate false positives and false negatives (inversely proportional) by treating the segmentation of each class as a binary classification problem. In our application, we note that the significance of the false positives and false negatives varies per class, as elaborated further in the upcoming section. However, we note in advance that we aim for a reasonably balanced trade-off between false positives and false negatives across sea, land, and cloud classes before selecting the final trained model. Therefore, we consider false positives and false negatives intuitively during our training procedure, but we acknowledge more rigorous methods that focus on minimizing either false positives or false negatives by adjusting decision thresholds. The adjustments can involve, for instance, altering the decision boundary during the categorical class assignment based on different probability thresholds, consequently impacting the false positives and false negatives for each class. Additionally, applying penalties within the loss function to address incorrect class predictions can also affect false positives and false negatives. However, in our work, we consider sufficient to predict the categorical classes during inference by simply selecting the class with the highest probability, while equally penalizing all classes during training as the dataset imbalance is moderate. 

In the supplementary material of this article, we provide the trained model files with the learned parameters (46 distinct files) for future in-orbit deployment. Additionally, we include a software data product with embedded inference by 1D-Justo-LiuNet within a user-friendly GUI for conducting oceanic-terrestrial-cloud segmentation on HYPSO-1 imagery.

        \subsection{Segmentation Performance Metrics and \\Proposed Example Application}
        \label{Section:ranking_application_case}

        \subsubsection{Proposed On-Board Sea-Land-Cloud Ranking System}

\begin{figure}[t]
  \centering
  \includegraphics[width=\columnwidth]{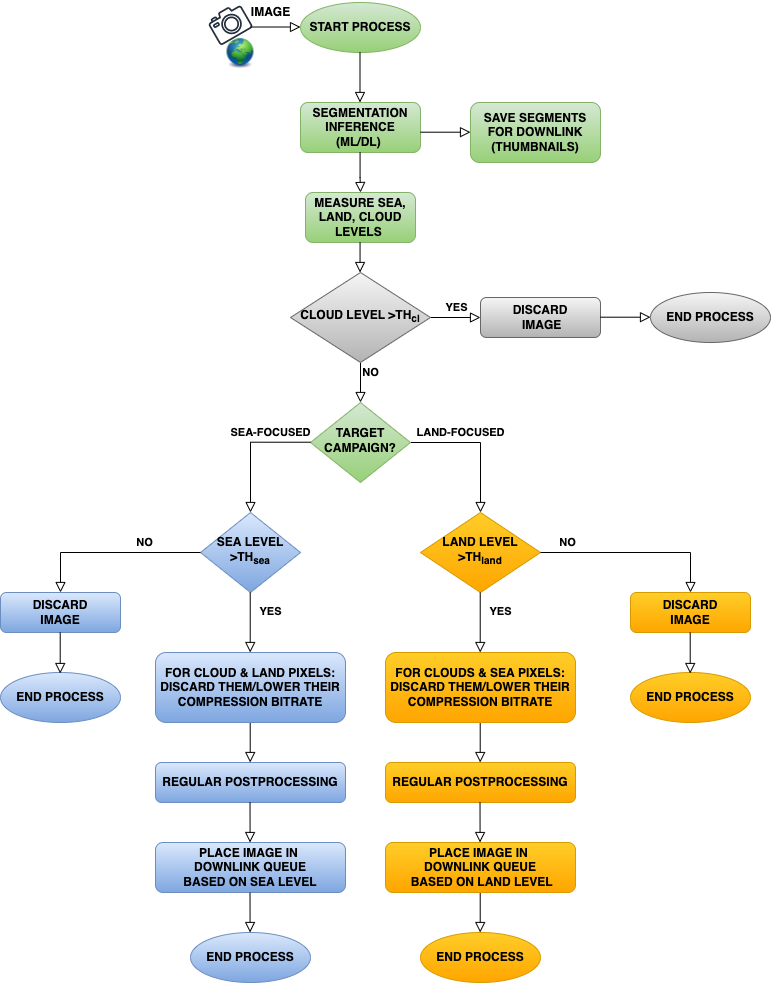}
  \caption{An automated on-board processing pipeline integrating a ranking system based on sea-land-cloud segmentation. $T\!H_{cl}$, $T\!H_{sea}$, and $T\!H_{land}$ represent thresholds configured at the ground center to regulate data management during in-flight operations.}
  \label{Fig:flow_chart_ranking_system}
\end{figure}

Before introducing the metrics to evaluate the segmentation outcomes, we next introduce example applications to study the utility of the sea, land, and cloud segments in increasing the satellite's autonomy, while optimizing operations, and reducing decision-making latency. Several performance metrics discussed in the following section are based on the next proposed example applications. In Fig.~\ref{Fig:flow_chart_ranking_system}, we suggest a novel sea-land-cloud ranking system for in-orbit deployment. The system prioritizes HS images for downlink based on sea, land, and cloud coverage levels, ensuring high-priority transmission. The images are arranged in ascending order of cloud coverage levels, descending order of sea coverage levels, and also decreasing order of land coverage levels, with the first positions reserved for images of higher downlink priority. Unlike existing ranking systems that target only clouds \cite{salazar2022cloud}, our proposed approach is the first of its kind for HSI-equipped satellites, simultaneously locating and ranking sea, land, and clouds within priority queues. In Fig.~\ref{Fig:flow_chart_ranking_system}, we also suggest other on-board use cases that further highlight the usefulness of the sea-land-cloud segments. The use cases include alternatives such as discarding or applying selective lossy compression (e.g., via Compressive Sensing~\cite{justo2022comparative}) to non-essential pixels to optimize the utilization of the Shannon channel capacity, particularly vital for satellites such as HYPSO-1, which have constrained downlink bandwidth and limited line-of-sight (LOS) to ground stations \cite{Bakken2023}.

        \subsubsection{Performance Metrics}

Regarding performance metrics, we utilize a range of metrics to assess the quality of the segmented images on the test set (prior to generalization tests in the deployment set) in order to evaluate the models after training. The employed metrics, including accuracy \cite{salazar2022cloud}, alarm ratios \cite{giuffrida2020cloudscout, pitonak2022cloudsatnet}, precision \cite{salazar2022cloud,spiller2022wildfire}, recall and F-1 score \cite{salazar2022cloud, spiller2022wildfire}, are frequent in the field of on-board segmentation for satellite spectral missions, enabling direct comparisons. However, our primary focus is on average pixel-level accuracy, a widely used metric for evaluating segmentation models. Unlike overall accuracy, this average metric is particularly convenient for imbalanced test sets, providing a more realistic assessment. Furthermore, we concentrate on the false alarm ratios for each individual class. Moreover, to evaluate the predicted rankings for each coverage level, we use the Spearman's rank correlation coefficient, following the literature \cite{salazar2022cloud}. This coefficient measures the relationship between the predicted and ground-truth rankings, with values ranging from -1 to 1, representing a perfect match at the upper end. To maintain a focused analysis, we employ these concrete metrics to compare model performance. However, we note that in our supplementary material, we also provide additional metrics such as the precision, recall, F1 score, overall accuracy, and the Mean Absolute Error (MAE) for each coverage level, treating each level similar to a regression task.

Regarding alarm ratios in the context of the on-board data management proposed in Fig.~\ref{Fig:flow_chart_ranking_system}, the false alarm ratios due to misclassifications indicate either the potential loss of data or the transmission of irrelevant data during downlink, as explained next. In cloud segmentation, the True Positive Ratio (TPR) involves correctly detecting cloudy pixels, while the True Negative Ratio (TNR) relates to recognizing non-cloudy data with potential value. On the one hand, the False Positive Ratio (FPR) is crucial because it indicates misclassified pixels, wrongly identified as clouds but that actually correspond to sea or land data. Following Fig. \ref{Fig:flow_chart_ranking_system}, these missclassifications can result in the discard or selective lossy compression with lower bitrate of valuable data. On the other hand, the False Negative Ratio (FNR) is less critical than FPR, as it does not involve data loss or data degradation, but only a less efficient usage of the downlink channel transmitting undetected cloud pixels. In the context of sea segmentation, true positives signify correct identification of valuable sea data, while true negatives correctly recognize non-relevant data for potential deletion or compression. In sea-focused campaigns, FNR is critical, as it deals with misclassified sea-related pixels, potentially leading to loss of valuable sea data or its degradation. In contrast, FPR is of less significance since its impact lies merely in downlink efficiency with the transmission of irrelevant pixels (clouds or land), rather than causing any data loss. The same analysis and conclusion apply when examining the context of land segmentation.

In summary, without taking into account the earlier suggested reasonable compromise between false positives and false negatives, for both sea and land categories, FNR has priority over FPR due to valuable data loss or degradation in the case of high FNR. However, for cloud data, FPR is critical as it leads to irreversible data loss or lossy compression, with FNR affecting merely downlink efficiency. It is noted that data loss in any communication channel, including a satellite radio channel, is undoubtedly more critical than utilizing the channel less efficiently. This is because lost data leads to permanent loss of information, or irreversible degradation below Shannon’s compression entropy, while the inefficient use of the downlink impacts \textit{only} other channel parameters such as transmission speed, transmission power and radio spectra utilization. While these downlink parameters can certainly be bottlenecks for (near) real-time applications and impact the satellite’s power budget, they do not lead to a critical scenario of absolute irrevocable data loss. Therefore, in our work, parameters affecting the downlink efficiency are given less weight compared to the irreversible loss or degradation of valuable data.

\section{Results}
\label{Section:Results}

\subsection{Main Results}

We conduct model training using 112 channels of both raw sensor data and L1b calibrated radiance, and we discuss the segmentation of raw and L1b radiance in Section \ref{Section: Discussion}. For L1b radiance, we present the outcomes of the segmentation models in Table~\ref{Table:main_metrics_for_all_models_112_COLORS}, where we utilize bold font to mark our models. The table displays the performance of 1D ML, 1D-CNNs, 2D-CNNs, and FastViT for segmentation, with the primary metric being the average accuracy across the sea, land, and cloud classes. Additionally, to assess the sea-land-cloud segmentation performance for the proposed on-board ranking system, we also include the average Spearman's rank coefficient, in Table~\ref{Table:main_metrics_for_all_models_112_COLORS}, calculated across the respective rankings based on sea, land, and cloud coverage. Moreover, we also include information about the network's parameter count to offer insights of the potential memory demands for in-orbit deployment. In addition, we provide the average inference time per image measured in a testing setup on ground, on a NVIDIA RTX A4000 GPU (for DL), and on the AMD Ryzen 7 5800X CPU (for ML). From the numbers presented in Table \ref{Table:main_metrics_for_all_models_112_COLORS}, we sort the top 10 models in Table \ref{Table:models_sorted_112_COLORS} based on their respective metrics. In Table \ref{Table:models_sorted_112_COLORS}, we first sort the models solely based on their average accuracy, from highest to lowest, and subsequently, we sort them in the following columns, first by the Spearman's coefficient (highest to lowest), then by the number of parameters (lowest to highest), and finally by inference time (shorter to longer). Therefore, the first rows in the table represent the top models in the test set for each respective metric.




\begin{table*}[htbp] 
\centering
\caption{Metrics for segmentation models trained over 112 spectral channels.}

  \begin{tabular}{|c|l|c|c|c|c|} 
    \hline  
     Type & Model & Accuracy & Spearman's Coefficient & Parameter Count & Inference Time [ms] \\ \hline 
     
    \multirow{7}{*}{1D}    & 1D-ML/SGD &  0.91     &  0.90  & -  & 72 \\  
        & 1D-ML/NB  &  0.71     & 0.27  &  -  & 1,234\\  
        & 1D-ML/LDA &  0.89     & 0.93  & -  & 190 \\  
        & 1D-ML/QDA &  0.89     & 0.63  & -  & 2,024  \\

        & \textbf{1D-Justo-LiuNet} & \underline{\textbf{0.93}}      &  \underline{\textbf{1.00}}  & \underline{\textbf{4,563}} & \textbf{166} \\  
        & \textbf{1D-Justo-HuNet}  &   \textbf{0.91}     &  \underline{\textbf{1.00}}  & \textbf{9,543}  & \underline{\textbf{69}} \\  
        & \textbf{1D-Justo-LucasCNN} &   \textbf{0.91}    &  \textbf{0.90} & \textbf{25,323} & \textbf{142} \\ \hline

        \multirow{12}{*}{2D}    & 2D-NU-Net-mod &  0.80     &  0.83  & 32,340 & 364 \\       
         & 2D-NU-Net &   0.81    &  0.77 &  40,332   & 374 \\

         & 2D-CUNet  &   0.72    & 0.83  & 67,299  & 272 \\  
         & 2D-CUNet++ &  0.83     & 0.77  & 24,619  & 294 \\

         & 2D-CUNet Reduced &   0.76     & 0.90  & 22,019 &  270 \\  
         & 2D-CUNet++ Reduced &  0.90    &  0.83  & 12,379 &  294 \\  

         & 2D-UNet FAUBAI &   0.88     & 0.90  & 26,534,211   &  462 \\  
         & 2D-UNet FAUBAI Reduced &  0.85     & 0.83  &  1,956,835 & 388 \\ 

         & \textbf{2D-Justo-UNet-Simple} &    \textbf{0.92}   & \textbf{0.93}  & \textbf{7,641} & \textbf{318}  \\  
           
         & FastViT-T8 & 0.82 & - & 5,033,795  & 132 \\
         & FastViT-S12 & 0.90 & - & 10,242,819 & 156 \\ 
         & FastViT-MA36 & 0.88 & - & 44,647,755 & 288 \\ \hline
  \end{tabular}

  \label{Table:main_metrics_for_all_models_112_COLORS}

\end{table*}

\begin{table*}[tbph]
\centering
\caption{Ranking of top 10 segmentation models trained over 112 channels by various metrics, from best to worst.}

\label{Table:models_sorted_112_COLORS}

\begin{tabular}{|c|l|l|l|l|}
\hline
Ranking & Accuracy & Spearman's Coeff. & Parameter Count & Inference Time \\
\hline

\rowcolor{lightgray}
\textit{1}    & \textbf{1D-Justo-LiuNet} &  \textbf{1D-Justo-LiuNet} & \textbf{1D-Justo-LiuNet} & \textbf{1D-Justo-HuNet} \\

\textit{2}    & \textbf{2D-Justo-UNet-Simple}   &  \textbf{1D-Justo-HuNet} & \textbf{2D-Justo-UNet-Simple} &  1D-ML/SGD \\

\textit{3}    & 1D-ML/SGD    &  \textbf{2D-Justo-UNet-Simple} & \textbf{1D-Justo-HuNet} & FastViT-T8 \\

\textit{4}    & \textbf{1D-Justo-HuNet}   &  1D-ML/LDA & 2D-CUNet++ Reduced & \textbf{1D-Justo-LucasCNN} \\

\textit{5}    & \textbf{1D-Justo-LucasCNN}    &  \textbf{1D-Justo-LucasCNN} & 2D-CUNet Reduced & FastViT-S12 \\

\textit{6}    & 2D-CUNet++ Reduced    &  1D-ML/SGD & 2D-CUNet++ & \textbf{1D-Justo-LiuNet} \\

\textit{7}    & FastViT-S12    &  2D-CUNet Reduced & \textbf{1D-Justo-LucasCNN} & 1D-ML/LDA \\

\textit{8}    & 1D-ML/LDA    &  2D-UNet FAUBAI & 2D-NU-Net-mod &  2D-CUNet Reduced\\

\textit{9}    & 1D-ML/QDA    &  2D-CUNet & 2D-NU-Net &  2D-CUNet \\

\textit{10}   & 2D-UNet FAUBAI   &  2D-NU-Net-mod  & 2D-CUNet & FastViT-MA36 \\










\hline
\end{tabular}

\end{table*}

\begin{figure*}[tbph]
  \centering
  \resizebox{2.26\columnwidth}{!}{\includegraphics{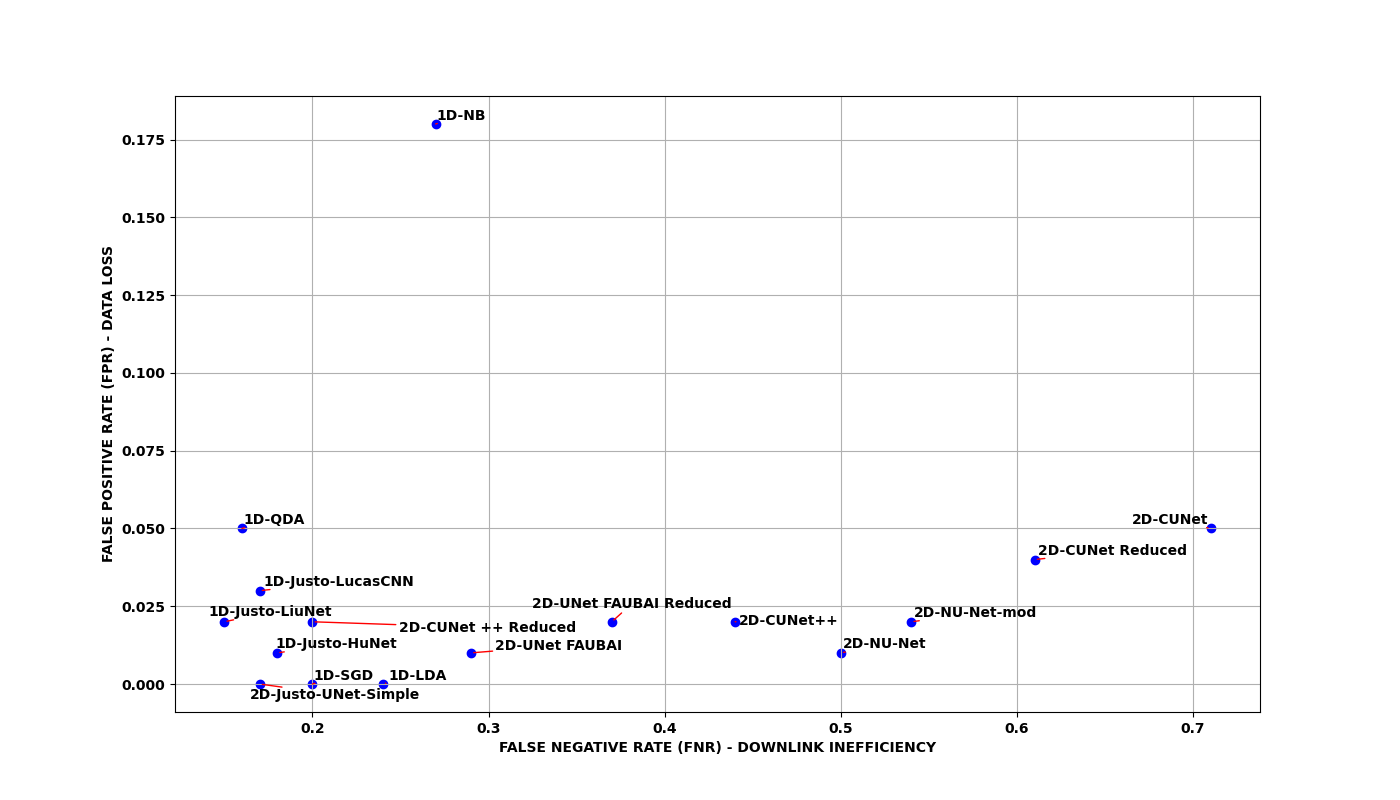}}

  \caption{Comparison of false detection ratios for the cloud class by segmentation models trained over 112 channels.}
\label{Fig:LOSS_OF_DATA_AND_INEFFICIENCY_FOR_CLOUDS_CLASS}
\end{figure*}

\begin{figure*}[tbph]
  \centering
    \resizebox{2.26\columnwidth}{!}{\includegraphics{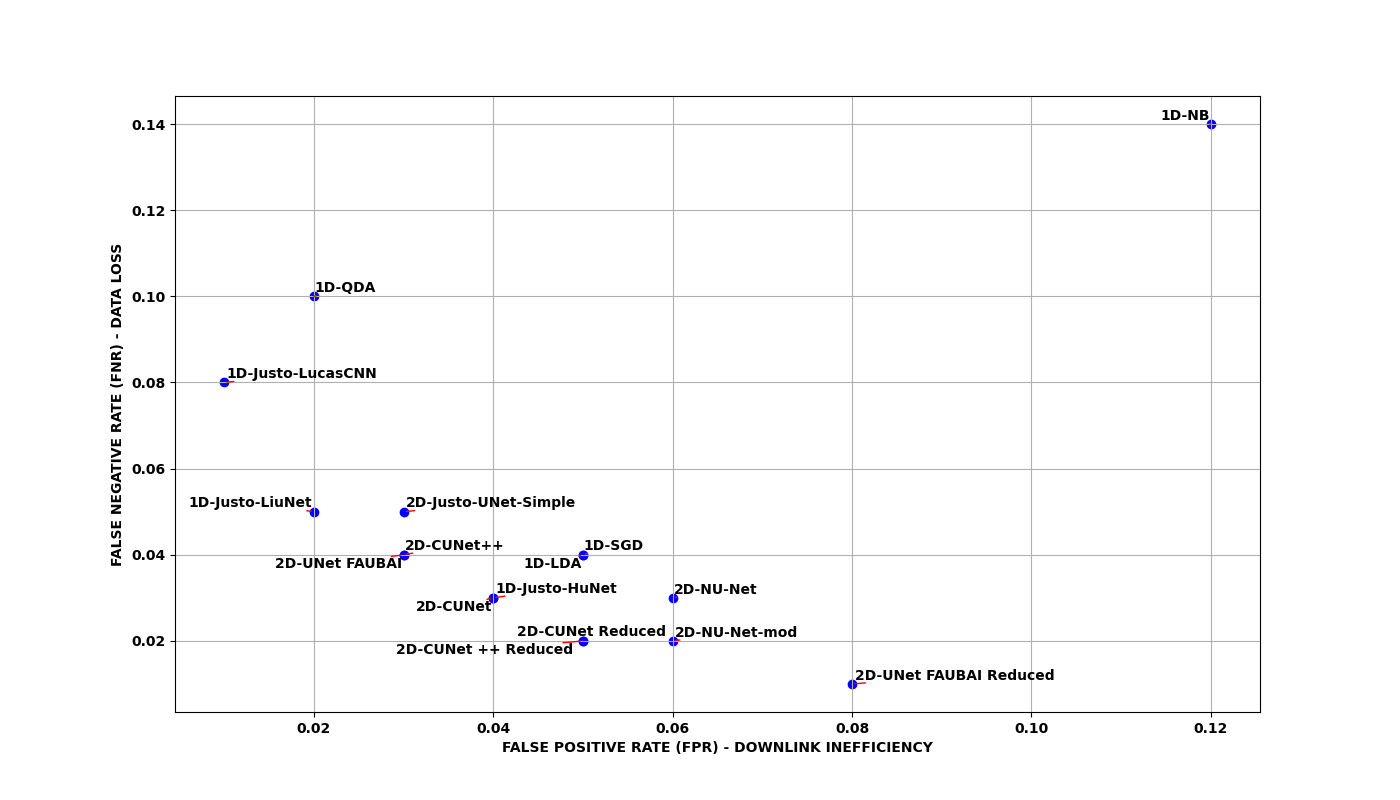}}

  \caption{Comparison of false detection ratios for the sea class by segmentation models trained over 112 channels.}
  \label{Fig:LOSS_OF_DATA_AND_INEFFICIENCY_FOR_SEA_CLASS}
\end{figure*}

\begin{figure*}[tbph]
  \centering
  \resizebox{2.2\columnwidth}{!}{\includegraphics{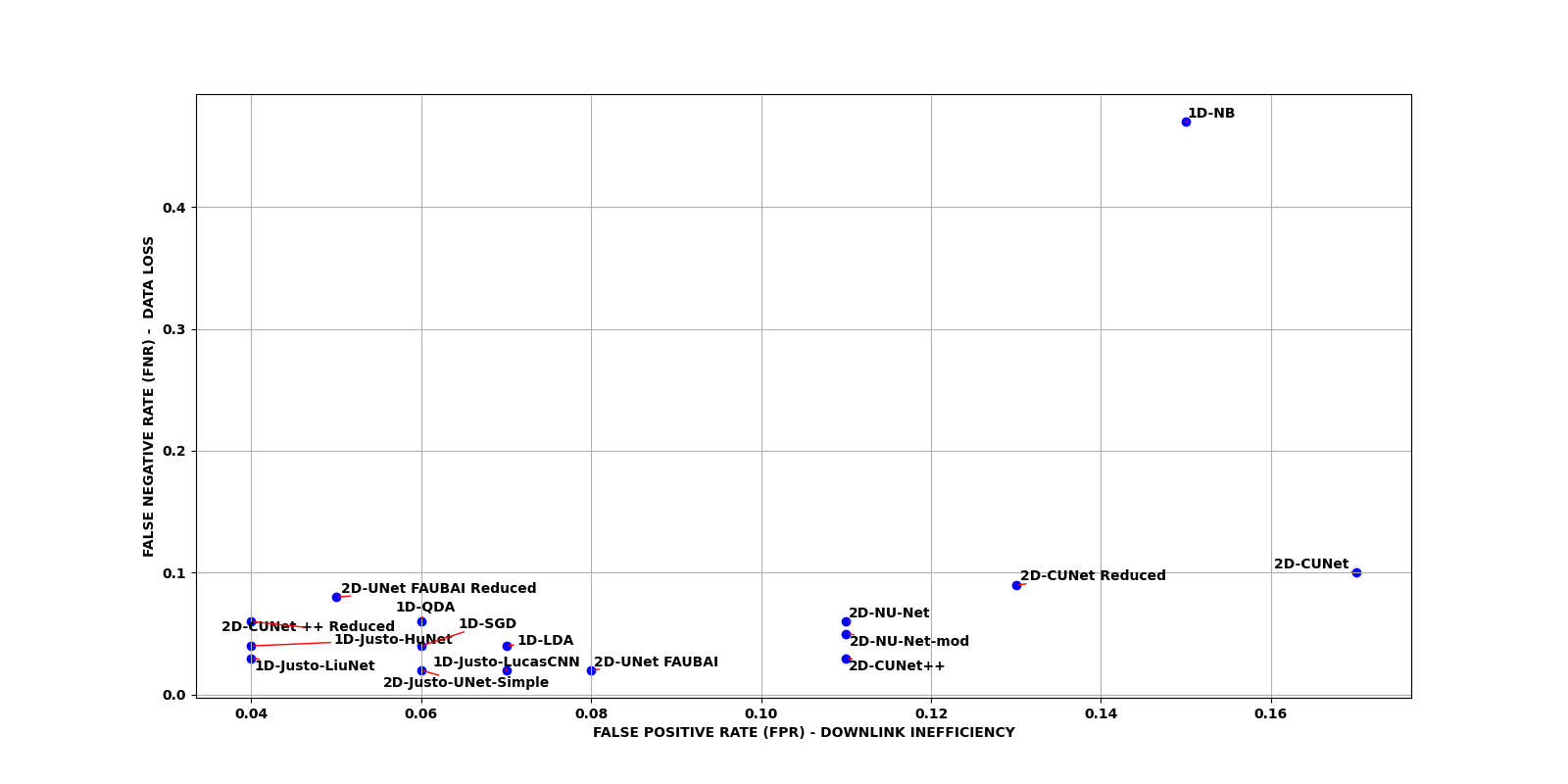}}

  \caption{Comparison of false detection ratios for the land class by segmentation models trained over 112 channels.}
  \label{Fig:LOSS_OF_DATA_AND_INEFFICIENCY_FOR_LAND_CLASS}
\end{figure*}

While average accuracy offers a general overview on segmentation performance across the sea, land, and cloud classes, it may obscure poor model performance in specific classes. To address this, we evaluate individual class performance by presenting their respective binary false detection ratios. Namely, in Figs.~\ref{Fig:LOSS_OF_DATA_AND_INEFFICIENCY_FOR_CLOUDS_CLASS}-\ref{Fig:LOSS_OF_DATA_AND_INEFFICIENCY_FOR_LAND_CLASS}, we provide a comparative visual representation of false alarm ratios for the binary segmentation of each distinct class. We note that the top-performing models get positioned in the lower-left corners, indicating the optimal trade-off between low false positives and low false negatives. Assuming the case example of the on-board ranking system, the graphs illustrate on the Y-axis the critical false alarm ratio corresponding to the loss (or degradation) of valuable data, while the X-axis shows the false alarm ratio that reflects the \textit{inefficient} utilization of the downlink channel. Both target variables should be minimized. As previously described, data loss is measured as FNR for sea and land categories and as FPR for clouds. Additionally, the \textit{inefficient} utilization of the channel is measured as FPR for sea and land categories and as FNR for clouds. This clarifies that in Fig.~\ref{Fig:LOSS_OF_DATA_AND_INEFFICIENCY_FOR_CLOUDS_CLASS} for clouds, the Y axis represents FPR, and the X axis represents FNR. In Figs.~\ref{Fig:LOSS_OF_DATA_AND_INEFFICIENCY_FOR_SEA_CLASS}-\ref{Fig:LOSS_OF_DATA_AND_INEFFICIENCY_FOR_LAND_CLASS} for sea and land, the Y axis represents FNR, and the X axis represents FPR. In relation to this trade-off, we stress that while minimizing the loss of valuable data remains crucial, we also highlight the importance of maintaining an acceptable downlink efficiency.

Moreover, Table~\ref{Table:main_metrics_for_all_models_3_COLORS} shows the results after retraining various models using the 3-channel HYPSO-1 dataset. We employ this dataset to assess the impact of channel reduction on performance, parameter count, and inference time. In addition, the results in Table~\ref{Table:main_metrics_for_all_models_3_COLORS} are further confirmed by training once again all models with a distinct set of three channels different from the bands used in Section \ref{Section:Methodology_for_Dimensionality_Reduction}. The new results align with the previous findings and for simplicity, we include them in the supplementary material. 

\begin{table*}[htbp] 
\centering
\caption{Metrics for segmentation models trained using only 3 spectral channels.}

  \begin{tabular}{|l|c|c|c|c|} 
    \hline  
     Model & Accuracy & Spearman's Coefficient & Parameter Count & Inference Time [ms] \\ \hline 

         2D-NU-Net-mod &   0.66      & 0.37  & 32,231 & 224 \\       
         2D-NU-Net &    0.80     & 0.53  &  32,375 & 192\\

         2D-CUNet  &    0.86     & \underline{0.83}  & 59,451 & 162 \\  
         2D-CUNet++ &    0.84     & 0.6  &  16,771 & \underline{100} \\

         2D-CUNet Reduced &   \underline{0.87}       & 0.73  & 14,171 & 156 \\  
         2D-CUNet++ Reduced &   0.84      & \underline{0.83}  & 4,531 & 122 \\  

         2D-UNet FAUBAI &  \underline{0.87}     & 0.77  &  26,471,427  & 346 \\  
         2D-UNet FAUBAI Reduced &   0.85      & 0.57 & 1,941,139 & 230 \\  

         \textbf{2D-Justo-UNet-Simple} &   \textbf{0.85}      & \underline{\textbf{0.83}}  & \underline{\textbf{1,755}} & \textbf{112}  \\ \hline 

  \end{tabular}

  \label{Table:main_metrics_for_all_models_3_COLORS} 

\end{table*}

\begin{figure*}[htbp]
  \centering
    \resizebox{0.8\textwidth}{!}{
    \includegraphics{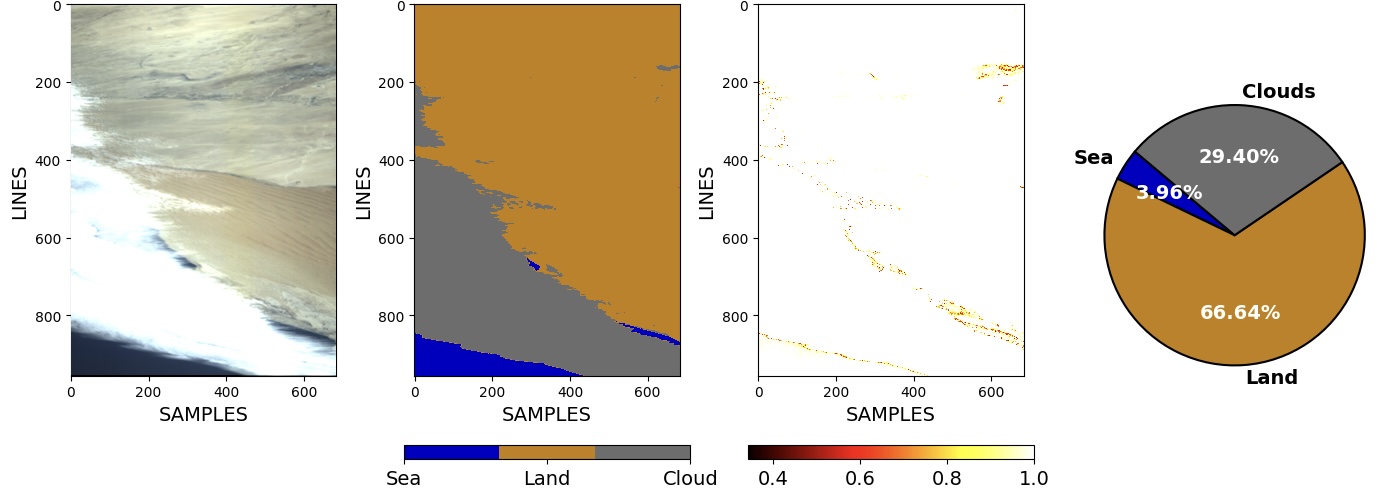}
    }
    \caption{Segmentation by 1D-Justo-LiuNet.}
\label{subfig:generalisation_test_1D_JUSTO_LIUNET}
\end{figure*}

\begin{figure*}[htbp]
    \centering
    \resizebox{0.8\textwidth}{!}{
    \includegraphics{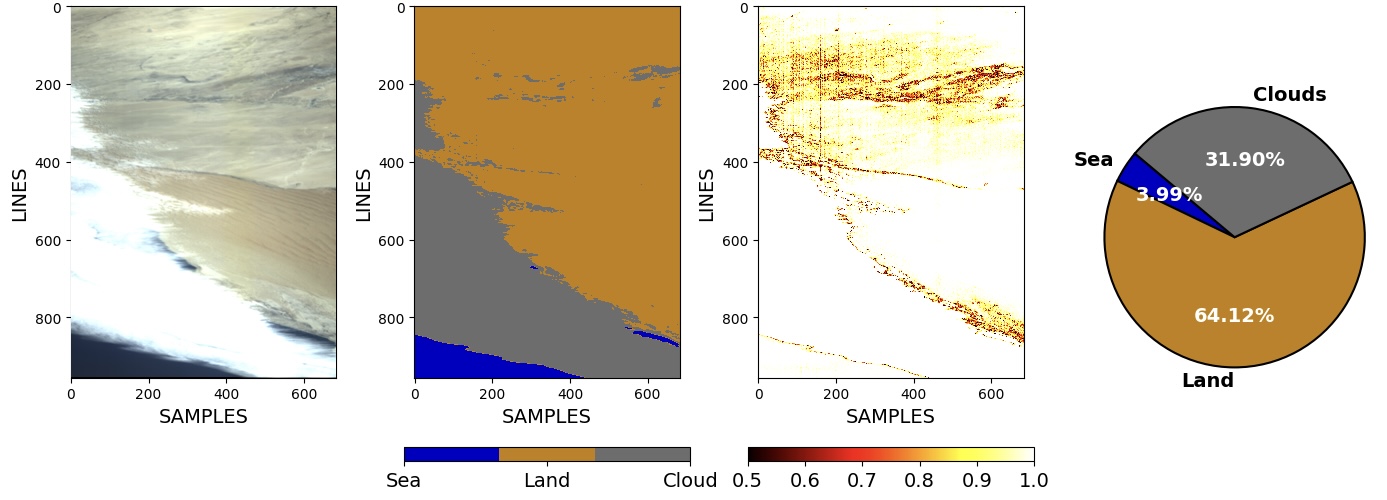}
    }
    \caption{Segmentation by 1D-Justo-HuNet.}
    \label{subfig:generalisation_test_1D_JUSTO_HUNET}
\end{figure*}

\begin{figure*}[htbp]
  \centering
        \resizebox{0.8\textwidth}{!}{
        \includegraphics{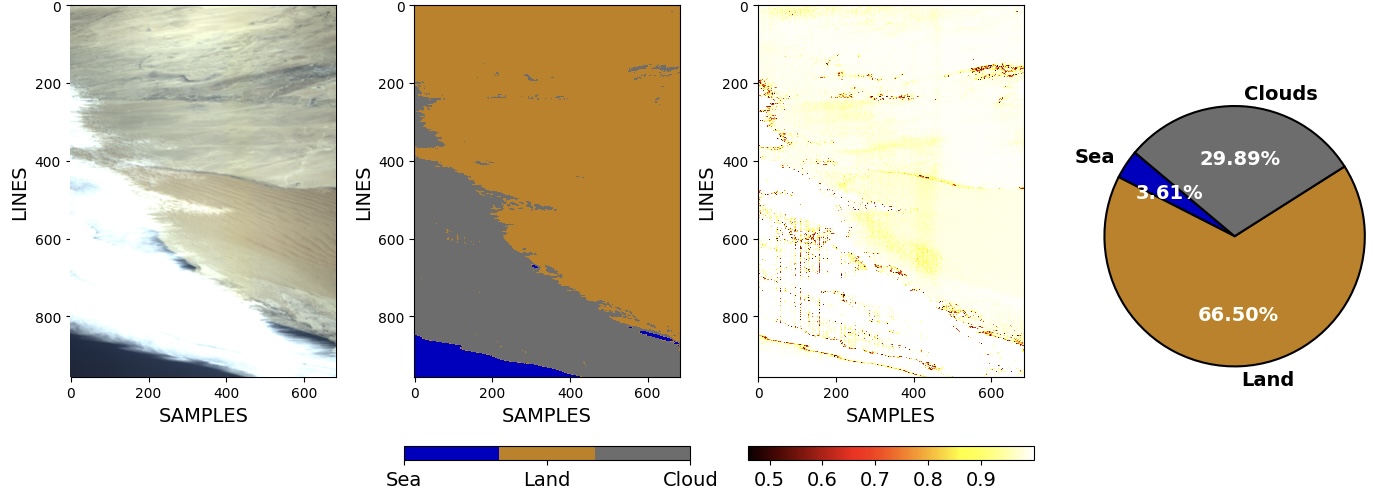}
    }
    \caption{Segmentation by 1D-Justo-LucasCNN.}
\label{subfig:generalisation_test_1D_JUSTO_LUCASCNN}

\end{figure*}

\begin{figure*}[htbp]
  \centering
    \resizebox{0.8\textwidth}{!}{
    \includegraphics{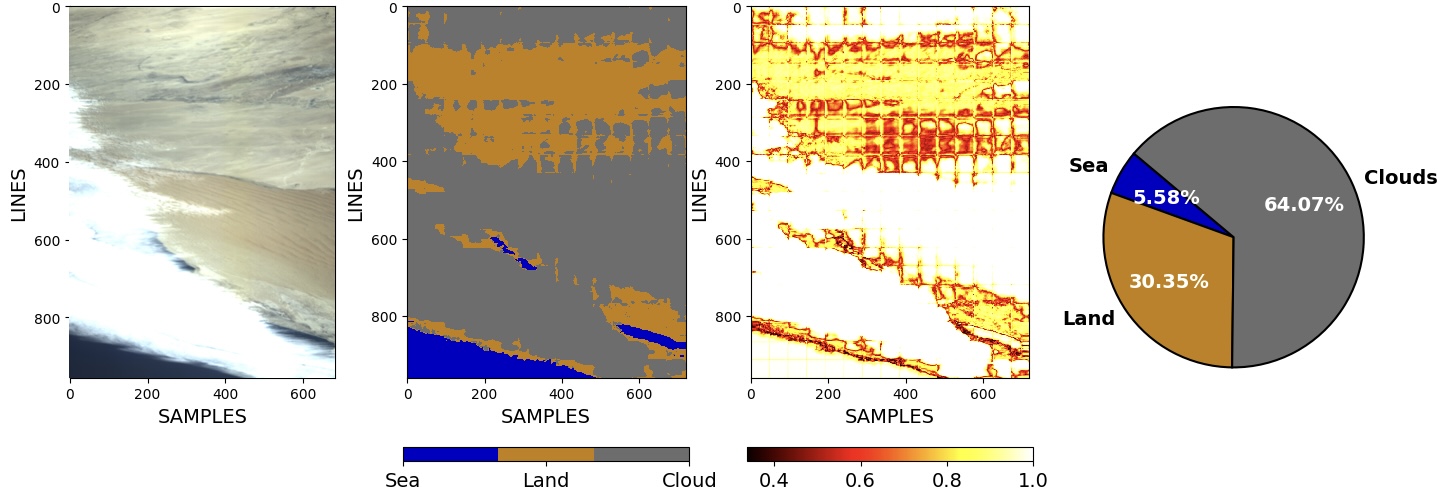}
    }
    \caption{Segmentation by 2D-NU-Net-mod.}
    \label{subfig:generalisation_test_2D_NUNETMOD}
\end{figure*}

\begin{figure*}[htbp]
  \centering
    \resizebox{0.8\textwidth}{!}{
    \includegraphics{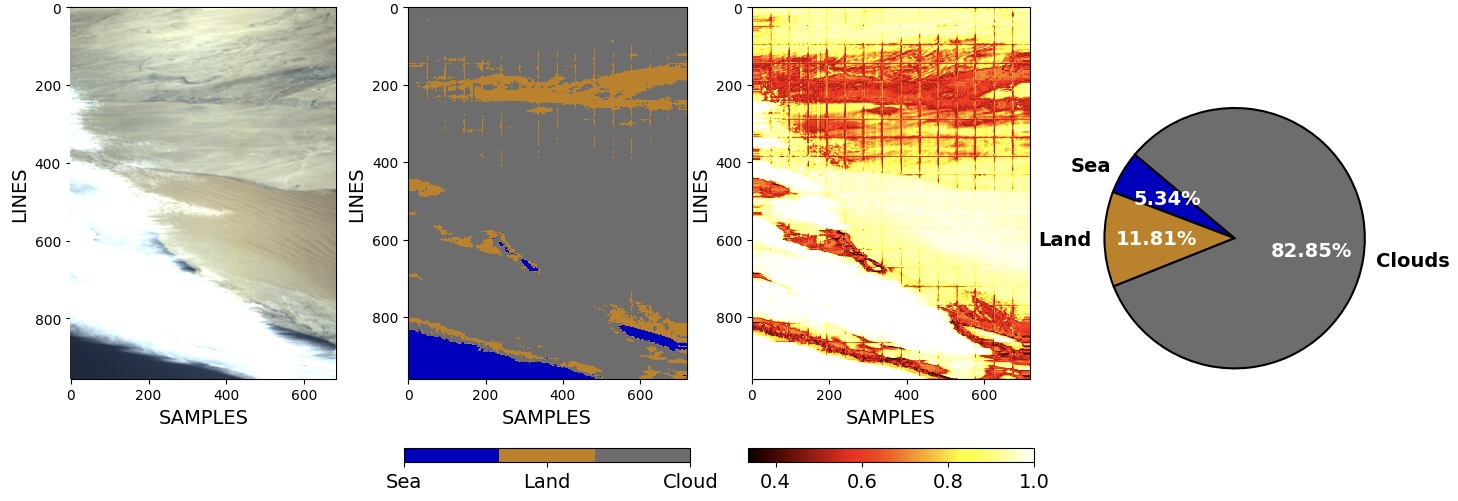}
    }
    \caption{Segmentation by 2D-NU-Net.}
    \label{subfig:generalisation_test_2D_NUNET}
\end{figure*}

\begin{figure*}[htbp]
  \centering
    \resizebox{0.8\textwidth}{!}{
    \includegraphics{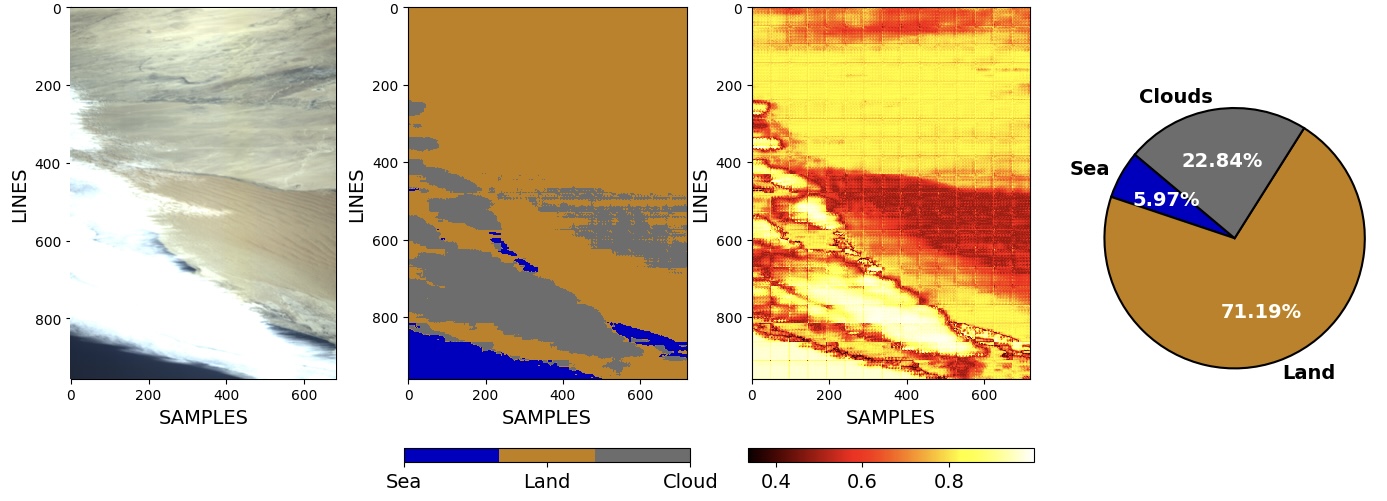}
    }
    \caption{Segmentation by 2D-CUNet.}
    \label{subfig:generalisation_test_2D_CUNET}
\end{figure*}

\begin{figure*}[htbp]
  \centering
    \resizebox{0.8\textwidth}{!}{
    \includegraphics{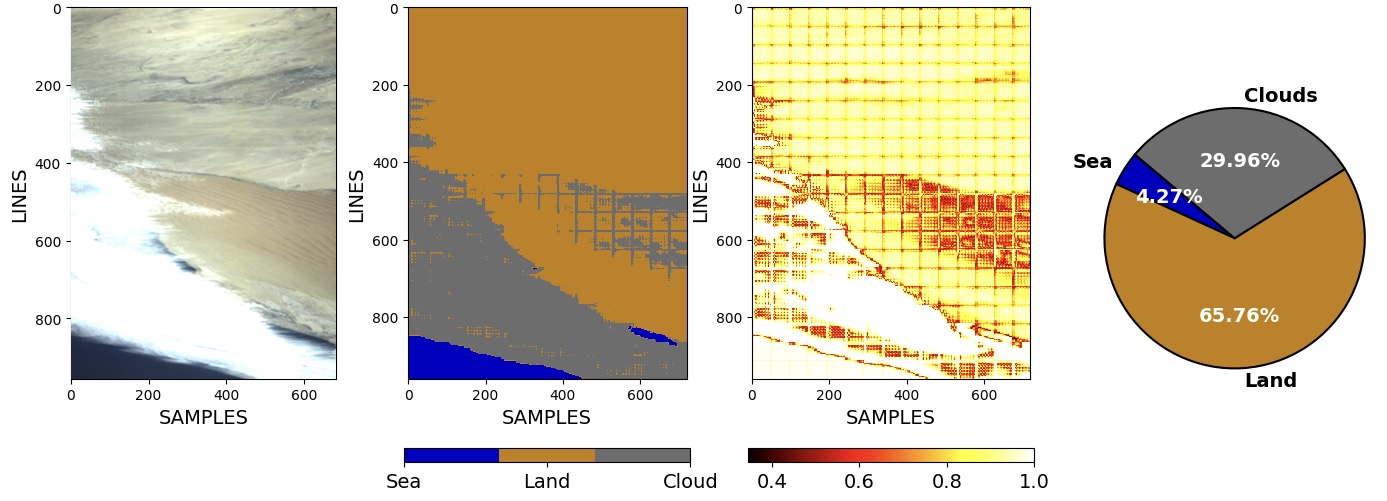}
    }
    \caption{Segmentation by 2D-CUNet++.}
    \label{subfig:generalisation_test_2D_CUNET++}
\end{figure*}

\begin{figure*}[htbp]
  \centering
    \resizebox{0.8\textwidth}{!}{
    \includegraphics{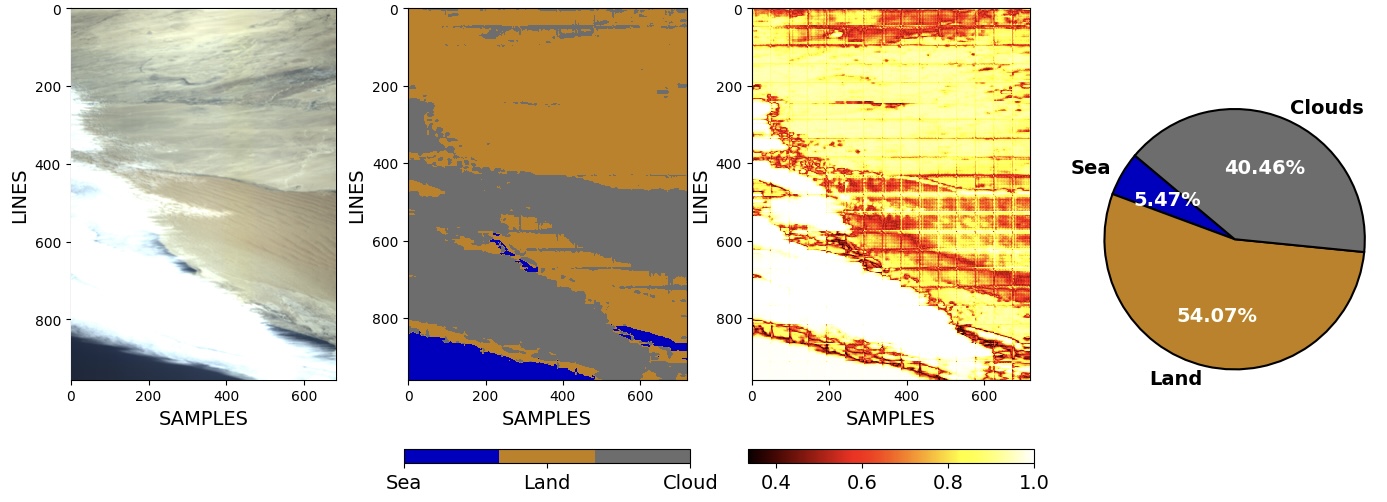}
    }
    \caption{Segmentation by 2D-CUNet Reduced.}
    \label{subfig:generalisation_test_2D_CUNET_Reduced}
\end{figure*}

\begin{figure*}[htbp]
  \centering
    \resizebox{0.8\textwidth}{!}{
    \includegraphics{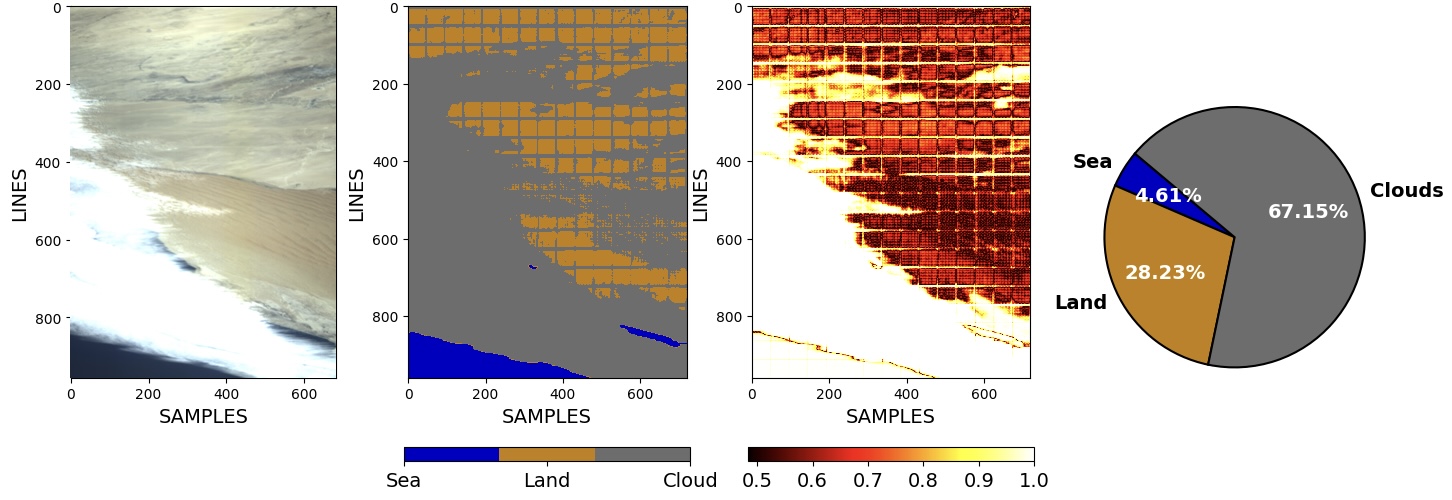}
    }
    \caption{Segmentation by 2D-CUNet++ Reduced.}
    \label{subfig:generalisation_test_2D_CUNET++_Reduced}
\end{figure*}

\begin{figure*}[htbp]
  \centering
    \resizebox{0.8\textwidth}{!}{
    \includegraphics{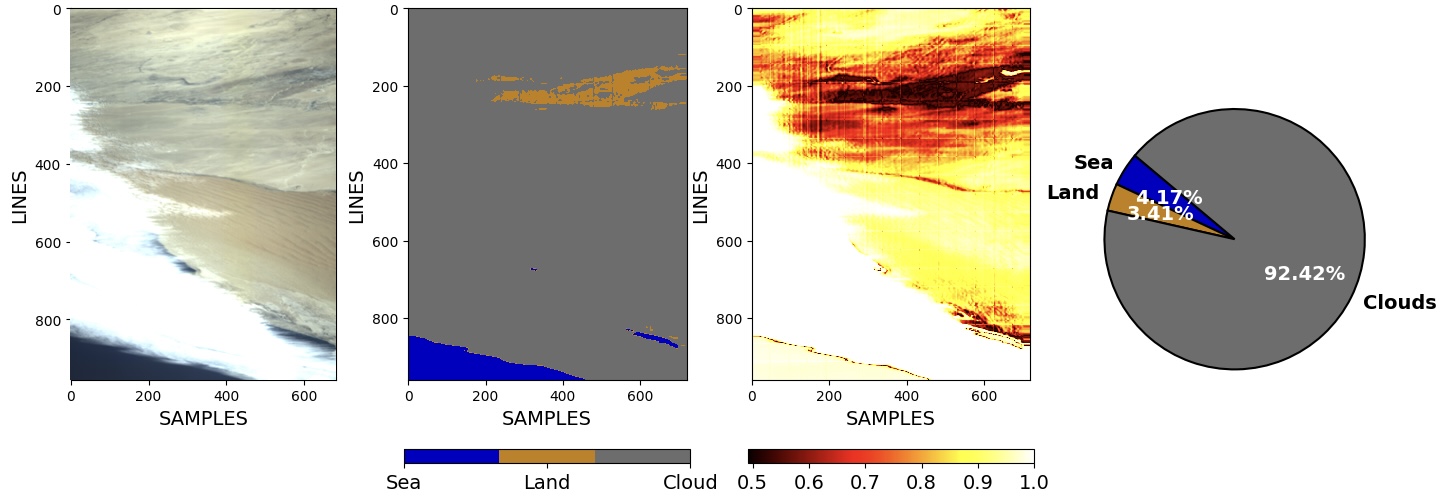}
    }
    \caption{Segmentation by 2D-UNet FAUBAI.}
    \label{subfig:generalisation_test_2D_UNET_FAUBAI}
\end{figure*}

\begin{figure*}[htbp]
  \centering
    \resizebox{0.8\textwidth}{!}{
    \includegraphics{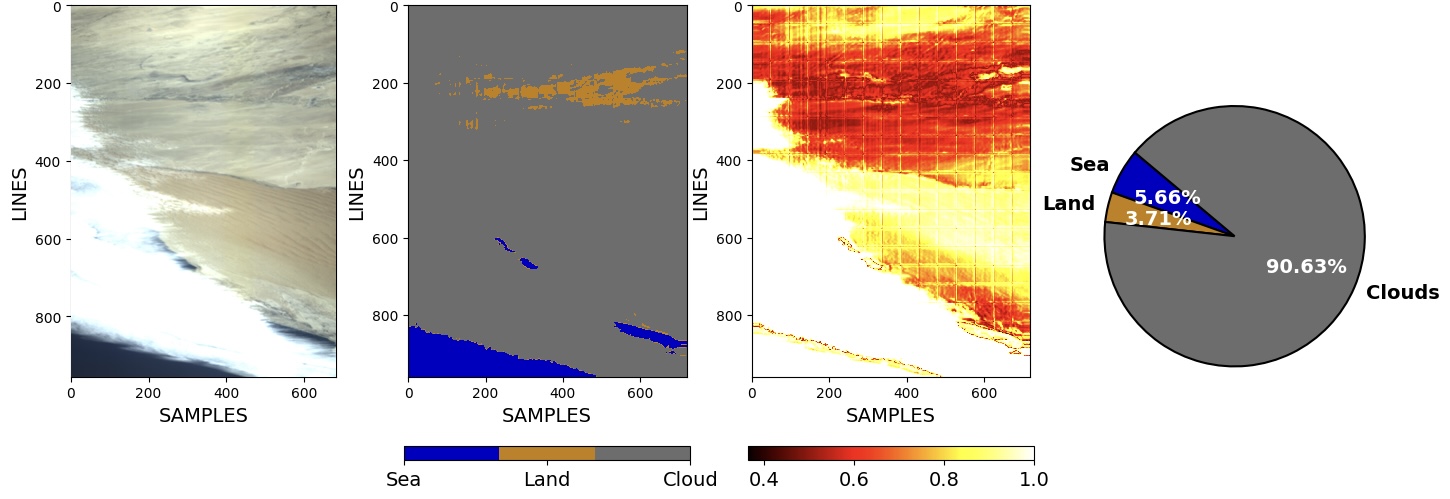}
    }
    \caption{Segmentation by 2D-UNet FAUBAI Reduced.}
    \label{subfig:generalisation_test_2D_UNET_FAUBAI_REDUCED}
\end{figure*}

\begin{figure*}[htbp]
  \centering
    \resizebox{0.8\textwidth}{!}{
    \includegraphics{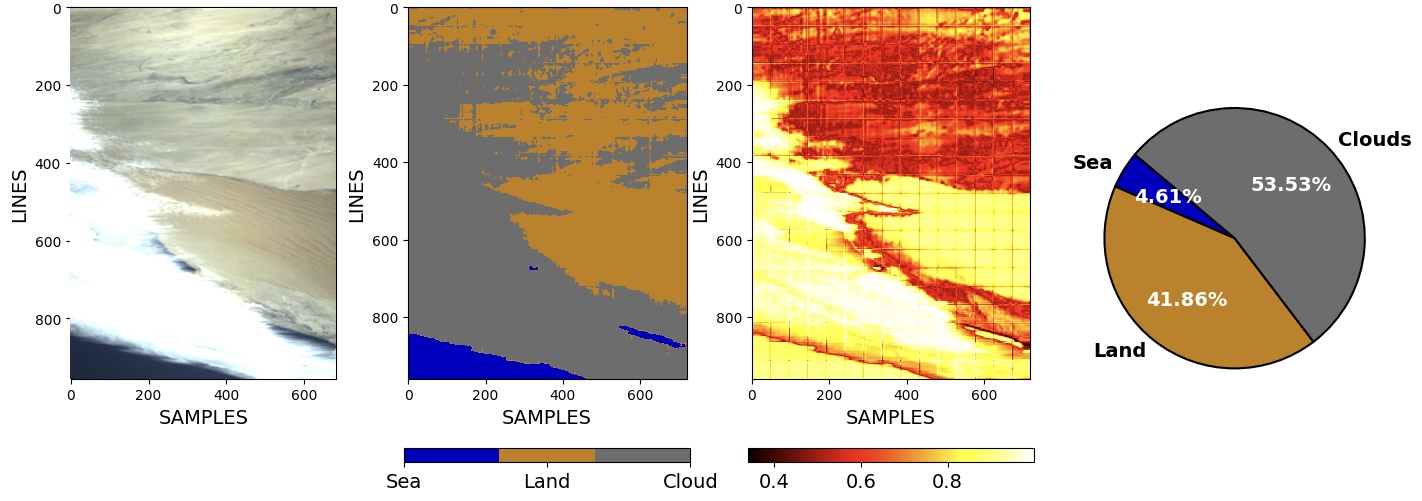}
    }
    \caption{Segmentation by 2D-Justo-UNet-Simple.}
    \label{subfig:generalisation_test_2D_JUSTO_UNET_SIMPLE}

\end{figure*}

As previously noted, the HYPSO-1 dataset from~\cite{justo2023open} predominantly contains instances of sea and land, with clouds representing the minority class. Consequently, the test set also exhibits a lower proportion of cloud data (14.13\%), in contrast to sea (39.01\%) and land (46.85\%). We recognize this as a limitation that may demand future refinement. To mitigate the impact of class imbalance during the testing phase, we conduct additional experiments. Namely, aligning with the segmentation literature for satellite imagery \cite{salazar2022cloud}, we perform inference on an additional on-ground deployment set, consisting of 30 HS images that are unseen by the model, to further test the model's generalization ability for HYPSO-1. We compare the performance of 1D models with 2D models on this deployment set, including cloudy and overexposed images. As an example, Figs.~\ref{subfig:generalisation_test_1D_JUSTO_LIUNET}-\ref{subfig:generalisation_test_2D_JUSTO_UNET_SIMPLE} depict how 1D models generalize compared with 2D models on a sample image from the deployment set. This image was captured by HYPSO-1 on 20 October 2022, over the Namib desert in Namibia, Africa, near the Gobabeb Namib Research Institute for dry land research~\cite{justo2023open}. The approximate coordinates of this location are 23°19'06.0"S latitude and 14°44'21.8"E longitude. For each figure, we present, from left to right: the RGB composite of the image captured by the satellite; the segmentation results; the model's confidence for each pixel-level prediction measured in terms of probability, with the highest confidence indicated by probabilities near the ideal value of 1.0 (100\%), representing no uncertainty; and finally, a pie chart showing coverage levels for the sea, land, and cloud categories to sort the images in the ranking system. In the figures, the term SAMPLES denotes the swath width captured across the direction of satellite travel, while LINES indicates the number of scan frames collected along the satellite's path. Finally, we note that the supplementary material includes further results for the remaining images in the deployment set, with inference run by 1D-Justo-LiuNet and 2D-CUNet++ Reduced.


\subsection{Generalization Results}

Despite our previous experiments on HYPSO-1 data used L1b calibration, a universal satellite data format, we next conduct additional tests on other datasets to evaluate generalization and validate our upcoming conclusions. Table~\ref{Table:generalisation_test_metrics_NASA_HYPERSPECTRAL} presents our results on NASA's EO-1 HS dataset for segmenting image pixels under three classes: \textit{no cloud}, \textit{thin cloud}, and \textit{thick cloud}. As we will discuss, the 1D-CNN 1D-Justo-LiuNet outperforms the state of the art on HS data. To verify this, we compare the top-performing 1D-Justo-LiuNet with a high-performing 2D-CNN in terms of accuracy and Dice coefficient. Finally, we also conduct experiments on the RGB Satellite Imagery of Dubai (SID) dataset. Table~\ref{Table:generalisation_test_metrics_RGB_AERIAL} presents the results for pixel-level segmentation across six classes: \textit{buildings}, \textit{roads}, \textit{land}, \textit{vegetation}, \textit{water}, and \textit{other}. The results include all 19 models previously tested for HYPSO-1.

\begin{table}[htbp] 
\centering
\caption{Metrics comparing top-performing 1D-CNN, 1D-Justo-LiuNet, with a high-performing 2D-CNN, tested on NASA's EO-1 hyperspectral dataset for generalization.}

  \begin{tabular}{|c|l|c|c|} 
    \hline  
     Channels & {Model} & {Accuracy} & {F1 Score} \\ \hline

      \multirow{3}{*}{98}  & \textbf{1D-Justo-LiuNet} & \underline{\textbf{0.82}} & \underline{\textbf{0.97}} \\  

        & nnU-net 2D & 0.81 & 0.87 \\   
        & \textbf{2D-Justo-UNet-Simple} & \textbf{0.70} & \textbf{0.81} \\ 
        
        \hline

      \multirow{3}{*}{6}  & \textbf{1D-Justo-LiuNet} & \textbf{0.74} & \textbf{0.85} \\  

        & nnU-net 2D & \underline{0.80} & 0.85 \\   
        & \textbf{2D-Justo-UNet-Simple} & \textbf{0.72} & \underline{\textbf{0.86}} \\ 
        
        \hline

      \multirow{3}{*}{1}  & \textbf{1D-Justo-LiuNet} & \textbf{0.64} & \textbf{0.76} \\  

        & nnU-net 2D & \underline{0.77} & 0.86 \\   
        & \textbf{2D-Justo-UNet-Simple} & \textbf{0.57} & \underline{\textbf{0.88}} \\ 
        
        \hline   
  
  \end{tabular}
  \label{Table:generalisation_test_metrics_NASA_HYPERSPECTRAL} 

\end{table}

\begin{table}[tbph] 
\centering
\caption{Metrics comparing 1D ML, 1D-CNNs, 2D-CNNs and ViTs models when tested on RGB Satellite Imagery of Dubai (SID) dataset for generalization.}

  \begin{tabular}{|c|l|c|c|} 
    \hline  
     Type & Model & Accuracy & Dice Coefficient \\ \hline 
     
    \multirow{7}{*}{1D}    & 1D-ML/SGD &  0.54     &  0.47  \\  
        & 1D-ML/NB  &  0.59     & 0.50  \\  
        & 1D-ML/LDA &  0.64     & 0.53  \\  
        & 1D-ML/QDA &  0.65     & 0.54  \\

        & \textbf{1D-Justo-LiuNet} & \textbf{0.63}      &  \textbf{0.51}  \\  
        & \textbf{1D-Justo-HuNet}  &   \textbf{0.67}     &  \textbf{0.58}  \\  
        & \textbf{1D-Justo-LucasCNN} &   \textbf{0.68}    &  \textbf{0.58} \\ \hline

        \multirow{12}{*}{2D}    & 2D-NU-Net-mod &  0.75    &  0.65  \\       
         & 2D-NU-Net &   0.78    &  0.69 \\

         & 2D-CUNet  &   0.77   & 0.68  \\  
         & 2D-CUNet++ &  0.77     & 0.67  \\

         & 2D-CUNet Reduced &   0.75     & 0.64  \\  
         & 2D-CUNet++ Reduced &  0.75    &  0.65 \\  

         & 2D-UNet FAUBAI &  0.82    & 0.70  \\  
         & 2D-UNet FAUBAI Reduced &  \underline{0.83}     & \underline{0.72}  \\ 

         & \textbf{2D-Justo-UNet-Simple} &   \textbf{0.76}   & \textbf{0.66}  \\  
           
         & FastViT-T8 & 0.82 & 0.71 \\
         & FastViT-S12 & 0.82 & 0.70 \\ 
         & FastViT-MA36 & \underline{0.83} & \underline{0.72}  \\ \hline
  \end{tabular}

  \label{Table:generalisation_test_metrics_RGB_AERIAL} 
\end{table}

\section{Discussion}
\label{Section: Discussion}

\subsection{Discussion of Main Results}

        \subsubsection{Segmentation Performance: Average Accuracy}

When evaluating the segmentation performance based on the average accuracy across sea, land, and cloud classes, we observe that the proposed model, 1D-Justo-LiuNet, achieves the highest accuracy at 0.93, as shown in Table \ref{Table:main_metrics_for_all_models_112_COLORS}. Our model surpasses the remaining 18 state-of-the-art models, including 2D-CNNs, and outperforms all FastViT configurations, including its lightest configuration, FastViT-T8, and its heaviest version, FastViT-MA36. The result of a 1D model outperforming 2D models is not an exception, as Table \ref{Table:models_sorted_112_COLORS} confirms that four out of the top five models in terms of accuracy are 1D models. The only exception is our 2D model, 2D-Justo-UNet-Simple, which achieves an accuracy of 0.92, closely approaching the performance of the 1D-Justo-LiuNet. These results can be attributed to the simplicity of 1D-CNNs, which extract features solely from the spectral domain. With fewer parameters and lower complexity, the risk of overfitting is reduced. In contrast, 2D models incorporate spatial context, significantly increasing the number of parameters and complexity, which raises the likelihood of overfitting, especially when training data is limited, making them more prone to learning spurious features. 
In short, 1D-CNNs achieve superior results in HS segmentation due to their simplicity, aligning with the principle of Occam's razor, which favors the simplest model as the best solution. We will demonstrate later that this conclusion also generalizes to other hyperspectral datasets.

        \subsubsection{Segmentation Performance: Priority Rankings for Downlink (Case Example)}

In this work, we simulate on ground a ranking system for future on-board deployment. From the labels in the 5 test images, we calculate the proportions of sea, land, and cloud classes. We emulate a downlink prioritization system based on cloud coverage, sorting from lowest cloud coverage (higher priority) to highest (lower priority). This serves as the ground-truth ranking based on cloud coverage. Next, a given model segments the images, and from these segments, we calculate the predicted class proportions. Using these proportions, we generate a predicted ranking based on cloud coverage, starting from lower to higher. Ideally, the predicted ranking would match the ground-truth ranking. To assess this, we compare the two rankings using Spearman's correlation coefficient. This allows us to see how the given model performs for a case application where the segments serve to guide data downlink to Earth, optimizing satellite operations and automation.

Table~\ref{Table:main_metrics_for_all_models_112_COLORS} shows that 1D-Justo-LiuNet, along with 1D-Justo-HuNet, outperforms the state of the art by achieving the highest possible performance, with a Spearman's coefficient of 1.00, resulting in an error-free ranking. Consistent with the accuracy results, Table~\ref{Table:models_sorted_112_COLORS} further confirms that four out of the top five models, now in terms of Spearman's coefficient, are 1D models. As we will discuss in the next section, the table also highlights that the parameter count for the lightest FastViT-T8 is in the order of millions, which is significantly  less suitable for in-orbit deployment compared with the few thousand parameters required by 1D-Justo-LiuNet and 1D-Justo-HuNet. Due to the high memory usage for storing parameters, FastViTs are not a suitable option for in-orbit deployment. Therefore, we exclude FastViTs from the ranking experiment. Moreover, we emphasize that FastViTs cannot surpass the ideal Spearman's coefficient of 1.00, already achieved by 1D-Justo-LiuNet and 1D-Justo-HuNet.

        \subsubsection{Parameter Count}

Table~\ref{Table:main_metrics_for_all_models_112_COLORS} focuses on the parameter count for deep models, as these typically have larger number of parameters compared with conventional ML algorithms. Parameter count is crucial for inference at edge due to the limited memory resources available on board. Tables~\ref{Table:main_metrics_for_all_models_112_COLORS} and \ref{Table:models_sorted_112_COLORS} show that 1D-Justo-LiuNet, which achieves top segmentation results, also outperforms all other state-of-the-art deep models in terms of parameter count, making it the lightest deep model. Thus, 1D-Justo-LiuNet is a highly compressed deep model that not only maintains but exceeds the segmentation performance of both conventional and the latest state-of-the-art models. We conclude that our network is the most suitable option for future deployment aboard HYPSO-1. To this extent, as an illustrative case, HYPSO-1 uses a Zynq-7030 System-on-Chip (SoC) with a dual-core ARM Cortex-A9 CPU and a Kintex-7 FPGA for on-board data processing~\cite{langer2023robust}. The dual-core CPU, limited by power constraints, has only 32 KB of L1 data cache for fast memory access~\cite{xilinx2018zynq}, theoretically allowing up to 8,000 model parameters to be stored (assuming 4-byte encoding). This illustrates the limited memory resources available on board. In our case, 1D-Justo-LiuNet requires only a few kilobytes for parameter storage, whereas even the lightest FastViT-T8, with millions of parameters and a higher risk of overfitting, demands considerably larger memory - in the order of megabytes. This makes FastViTs more challenging for in-orbit deployment.

        \subsubsection{Inference Time}

Table~\ref{Table:main_metrics_for_all_models_112_COLORS} shows that 2D-CNNs take longer inference times compared with 1D-CNNs and ViTs. For this analysis, we focus on the lightest models: 1D-Justo-LiuNet, 1D-Justo-HuNet, and ViTs such as FastViT-T8 and FastViT-S12. Although FastViT-T8 has a faster inference time of 132 ms, 1D-Justo-LiuNet is only 34 ms slower at 166 ms. However, 1D-Justo-LiuNet achieves significantly higher accuracy at 0.93, compared with FastViT-T8's 0.82. Considering this accuracy boost and the fact that 1D-Justo-LiuNet has over 1,000 times fewer parameters, we conclude that our model turns out as the best-performing model for in-orbit deployment. Furthermore, FastViT-T8, despite its speed, is not the fastest model. Our model 1D-Justo-HuNet achieves the most optimal inference time at just 69 ms, which is 63 ms faster than FastViT-T8. With an accuracy of 0.91 and a slightly higher parameter count than 1D-Justo-LiuNet, 1D-Justo-HuNet still outperforms all FastViTs in terms of accuracy and parameter counts, while also surpassing their inference time. Indeed, 1D-Justo-HuNet surpasses all the remaining 18 state-of-the-art models. Therefore, it is also a very feasible alternative for in-orbit deployment.

        \subsubsection{Class Segmentation Performance: False Detection Ratios, Data Loss, and Downlink Inefficiency}

For a more in-depth assessment of the segmentation performance at the class level, we scrutinize which models are positioned in the lower-left corner in the scatter plots in Figs.~\ref{Fig:LOSS_OF_DATA_AND_INEFFICIENCY_FOR_CLOUDS_CLASS}-\ref{Fig:LOSS_OF_DATA_AND_INEFFICIENCY_FOR_LAND_CLASS}. Given our earlier discussion on FastViTs and their large number of parameters for in-orbit deployment, the figures instead focus on lighter ML and DL models. The lower-left region ensures the most optimal false detection ratios, and within the example ranking system, the region achieves minimal loss (or degradation) of valuable data, while also minimizing the \textit{inefficient} utilization of the satellite downlink. As mentioned, lower data loss (in the Y-axis) takes precedence, while the downlink \textit{inefficiency} (in the X-axis) is of lower priority since it does not result in any loss of valuable data. However, we still balance a reasonable relation between false positives and false negatives to prevent scenarios where a model achieves minimal data loss at the cost of a highly inefficient use of downlink resources.

Next, we comparatively analyze the top-performing models located in the lower-left corners of Figs.~\ref{Fig:LOSS_OF_DATA_AND_INEFFICIENCY_FOR_CLOUDS_CLASS}-\ref{Fig:LOSS_OF_DATA_AND_INEFFICIENCY_FOR_LAND_CLASS}. To begin, we assess the models only in terms of data loss before we include downlink efficiency. In Fig.~\ref{Fig:LOSS_OF_DATA_AND_INEFFICIENCY_FOR_CLOUDS_CLASS} for cloud segmentation, models such as 2D-Justo-UNet-Simple excels in minimizing the loss of data, with nearly no data loss (ideal case). It means that the 2D-Justo-UNet-Simple outperforms by not misclassifying valuable non-cloud data (sea or land) as clouds, even if it means some cloudy data remains undetected, thus reducing the downlink efficiency. With respect to sea segmentation (Fig.~\ref{Fig:LOSS_OF_DATA_AND_INEFFICIENCY_FOR_SEA_CLASS}), the reduced 2D-UNet FAUBAI stands out for its minimal loss of sea pixels with FNR=0.01, i.e., there is only a 1\% proportion of sea pixels that remain undetected, resulting in their loss. This outperforms other models such as the 1D-Justo-LiuNet with 5\% loss. Finally, in Fig.~\ref{Fig:LOSS_OF_DATA_AND_INEFFICIENCY_FOR_LAND_CLASS} for land segmentation, we observe a remarkable similarity in false alarm ratios in the lower-left optimal region, resulting in a slight overlap. Therefore, we next clarify that the optimal models, when sorted in ascending order of FNR (data loss), are as follows: 0.02 (2D-Justo-UNet-Simple), 0.03 (1D-Justo-LiuNet), 0.04 (1D-Justo-HuNet), and 0.06 (2D-CUNet++ Reduced). Consequently, the 2D-Justo-UNet-Simple is the most optimal at minimizing the loss of land data, achieving only a 2\% ratio of undetected land pixels. However, although data loss is certainly critical, this analysis does not consider the balance between data loss and downlink efficiency. Therefore, to jointly maintain a reasonable efficiency while achieving low data loss, we next study the models taking into account this trade-off.

We could consider the use of Pareto fronts, which are typically applied in multi-objective optimization to address trade-offs between inversely proportional variables, as is the case here. Nevertheless, for simplicity, we analyze the trade-off using the Euclidean distance, which gives equal weight to the two objectives. Namely, we calculate the distance of each model from the origin point, representing the ideal scenario with zero false alarms. First, for cloud segmentation (Fig.~\ref{Fig:LOSS_OF_DATA_AND_INEFFICIENCY_FOR_CLOUDS_CLASS}), the models closest to the ideal false-free scenario have (FPR, FNR) values of (0.02, 0.15) for 1D-Justo-LiuNet and (0.00, 0.17) for 2D-Justo-UNet-Simple, resulting in approximate Euclidean distances of 0.151 and 0.170, respectively. Therefore, 1D-Justo-LiuNet exhibits the shortest distance to the ideal scenario. Second, the top models for sea segmentation (Fig.~\ref{Fig:LOSS_OF_DATA_AND_INEFFICIENCY_FOR_SEA_CLASS}) - with (FNR, FPR) of (0.08, 0.01) for 1D-Justo-LucasCNN, (0.05, 0.02) for 1D-Justo-LiuNet, and (0.04, 0.03) for 2D-UNet-FAUBAI - have approximate distances of 0.081, 0.054, and 0.050, respectively. This suggests a better performance for 2D-UNet-FAUBAI, with a marginal difference of 0.004 in distance compared with the 1D-Justo-LiuNet. However, Table \ref{Table:main_metrics_for_all_models_112_COLORS} notes that the 2D-UNet-FAUBAI has over 26 million parameters, whereas our network, 1D-Justo-LiuNet, has only 4,563 parameters, which is clearly more suitable for on-board deployment. Balancing the nearly identical performance and the large parameter difference against FAUBAI's network, 1D-Justo-LiuNet remains the top model for sea segmentation as well. Finally, for land segmentation (Fig.~\ref{Fig:LOSS_OF_DATA_AND_INEFFICIENCY_FOR_LAND_CLASS}), the optimal models are of (FNR, FPR) = (0.02, 0.06) for 2D-Justo-UNet-Simple, (0.03, 0.04) for 1D-Justo-LiuNet, (0.04, 0.04) for 1D-Justo-HuNet, and (0.06, 0.04) for 2D-CUNet++ Reduced, with resulting distances of 0.063, 0.050, 0.057, and 0.072, respectively. Therefore, 1D-Justo-LiuNet outperforms the other models for land segmentation.

In summary, when evaluating segmentation performance for each individual class considering a reasonable trade-off between data loss and downlink efficiency, 1D-Justo-LiuNet gnerally outperforms the other state-of-the-art models. We remark that 1D-Justo-LiuNet also excels in minimizing data loss, achieving acceptable low rates of 2\%, 5\%, and 3\%, while simultaneously demonstrating relatively minimal \textit{inefficiency} rates of 2\%, 3\%, and 15\% for sea, land, and cloud classes, respectively. On the one hand, these findings suggest that downlink efficiency slightly deteriorates due to an under-detection of clouds, leading to an overabundance of 15\% cloud data that would be downlinked from the satellite. Despite not critical, we attribute this to the dataset's class imbalance, where the cloud class constitutes the minority. On the other hand, the over-detection of the majority classes sea (2\%) and land (3\%) pixels is smaller, thereby increasing the overall efficiency of the downlink channel.

Before concluding the discussion on model performance, we compare our results with other state-of-the-art methods for HS segmentation. For example, a Support Vector Machines (SVM) model was previously tested and deployed in orbit for HS segmentation on HYPSO-1~\cite{roysland2023hyperspectral}. The SVM achieved a maximum average accuracy of 0.85, with a significantly lower minimum accuracy of 0.19. In contrast, 1D-Justo-LiuNet demonstrates a higher accuracy of 0.93, while the lowest accuracy, by 2D-CUNet, is 0.72, improving overall performance on HYPSO-1.

        \subsubsection{Generalization Performance in HYPSO-1 Deployment Set}

Up to this point, we have reported results based on the test set, where metrics such as accuracy and Spearman's coefficient could be calculated using ground-truth labels. However, to evaluate the model on a larger, unseen dataset, we utilize a deployment set, which  lacks ground-truth. As a result, the further generalization assessment in HYPSO-1 is conducted through expert visual inspection, a common practice in the literature~\cite{salazar2022cloud}. Given the promising results of lightweight 1D-CNNs for future edge inference, we focus on their generalization capabilities compared with 2D-CNNs, which have fewer parameters than ViTs, making them more suitable for on-board inference. When comparing the segmentation results of 1D-CNN models in Figs.~\ref{subfig:generalisation_test_1D_JUSTO_LIUNET}-\ref{subfig:generalisation_test_1D_JUSTO_LUCASCNN} with the results from 2D-CNNs in Figs.~\ref{subfig:generalisation_test_2D_NUNETMOD}-\ref{subfig:generalisation_test_2D_JUSTO_UNET_SIMPLE}, we confirm that 1D-CNNs outperform 2D-CNNs, with 1D-Justo-LiuNet achieving the best performance (Fig.~\ref{subfig:generalisation_test_1D_JUSTO_LIUNET}).

Regarding the sea, land, and cloud coverage levels shown in the pie charts in the same figures, a cloud coverage level of around 30\% is reasonable to expect for this scene. The 1D models predict this level accurately, while the 2D models often yield significantly different results, indicating much higher cloud coverage. This discrepancy can lead to the irrevocable loss (or degradation) of valuable surface data. Although 2D-CUNet (Fig.~\ref{subfig:generalisation_test_2D_CUNET}) and 2D-CUNet++ (Fig.~\ref{subfig:generalisation_test_2D_CUNET++}) may provide acceptable cloud coverage levels, they misclassify some ground data as clouds, still resulting in irrevocable loss of information. Thus, 1D-CNNs such as 1D-Justo-LiuNet remain the most suitable option in terms of performance. Even when considering the model's confidence in terms of probability shown in the figures, 1D models show more certainty in pixel-level predictions compared with 2D models. As a matter of fact, 1D-Justo-LiuNet (Fig.~\ref{subfig:generalisation_test_1D_JUSTO_LIUNET}), clearly stands out with the highest confidence and almost no uncertainty, whereas the remaining models, especially the 2D ones, exhibit significantly lower certainty.

Furthermore, we note that for all the models, feature extraction is challenging at class boundaries, with lower certainty at these transitions. This issue may increase the risk of higher misclassification, for instance, in images with high-entropic class distributions or scenes with high variability. We attribute this to the fact that 1D models do not consider spatial neighboring context, while 2D models may still require additional training and regularization to improve performance at class transitions. Additionally, we observe that 1D models generally produce predictions with higher entropy. This is because they lack spatial context unlike 2D models, leading to relatively less smooth transitions between predicted classes. Finally, Figs.~\ref{subfig:generalisation_test_2D_NUNETMOD}-\ref{subfig:generalisation_test_2D_JUSTO_UNET_SIMPLE} show that 2D models tend to misclassify pixels at the edges of the non-overlapping patches, exhibiting reduced certainty in these tile regions. This issue may arise from several factors, including the padding effect of the network on the edges of each input patch being processed. Based on the overall analysis of generalization in HYPSO-1, we conclude that 1D-Justo-LiuNet remains the top-performing model for future in-orbit deployment.

        \subsubsection{Models Trained on 3 Channels}

We also address the impact of reducing the number of channels from 112 to 3 on the reported metrics. As indicated in Table \ref{Table:main_metrics_for_all_models_3_COLORS}, when we train the models using the 3-channel dataset from HYPSO-1, the inference times decrease compared with the models trained with 112 channels, as there is now less data to process. Moreover, there is a reduction in the number of model parameters due to the smaller number of components within each convolutional kernel when computing the respective 2D convolutions across the reduced spectral dimension. However, the reduction in feature channels leads to a decreased segmentation performance, with the highest accuracy reaching 0.87, unlike the 0.93 top accuracy for the models trained with 112 channels. The decline in performance is also evident in the evaluation of the ranking system, for which none of the models achieve the ideal Spearman's coefficient of 1.00, as observed earlier with 112 channels. As a matter of fact, the highest coefficient for 3 channels is notably lower at 0.83. In conclusion, to achieve higher segmentation performance in HS imagery, inference should be conducted using most spectral channels due to the richness of information in this domain. Given the high dimensionality of the data, to reduce complexity, 1D-CNNs should be utilized, which also boost segmentation performance over 2D-CNNs. Although Table~\ref{Table:main_metrics_for_all_models_3_COLORS} does not include results for 1D-CNNs and FastViTs using only 3 channels, we perform these experiments on the RGB SID dataset.

        \subsubsection{Knowledge Distillation}

In an attempt to boost the performance of 1D-Justo-LiuNet, we conduct additional experiments using knowledge distillation, with FastViT-S12 as the teacher model and 1D-Justo-LiuNet as the student. However, we do not obtain any performance improvements, i.e., the accuracy of 1D-Justo-LiuNet remains at 93\%. This can be explained by the fact that, as shown in Table~\ref{Table:main_metrics_for_all_models_112_COLORS}, 1D-Justo-LiuNet has a higher performance level (0.93) than the FastViT-S12 teacher (0.90).

       \subsubsection{The Effect of L1b Radiance Calibration}

Before concluding the discussion of the main results, we address the impact of L1b radiance calibration on segmentation performance. Table \ref{Table:main_metrics_for_all_models_112_COLORS} shows models trained on L1b calibrated radiance, a standard format in HSI. Radiance not only allows satellite comparisons but also removes sensor-specific characteristics and biases. We trained our models in radiance to ensure that the results can be generalized to satellites different from HYPSO-1. Beyond this, it is of interest to assess how L1b radiance calibration affects segmentation performance to determine whether in-orbit inference should occur before or after calibration.

For brevity, we provide additional comparative tables in the supplementary material showing segmentation results when models are trained for pre- and post-calibration inference. The highest-performing 1D-CNN models show slightly better performance with L1b radiance than with raw data. A similar trend is observed with top 2D-CNNs, which also perform marginally better with radiance. However, lower-performing models tend to perform better with raw data. While the top-performing models benefit slightly from radiance calibration, the improvement is minimal and may not justify the computational cost of performing calibration in orbit. In any case, we attribute the marginal improvement to calibration correcting errors and biases, allowing top-performing models to focus on inherent data patterns rather than overfitting on sensor-induced characteristics, thus improving generalization. However, we recommend testing both calibrated and raw sensor data when training models on different datasets to verify if calibrated data, such as radiance or reflectance, continues to enhance performance.

\subsection{Discussion of Generalization Results}

To validate the generalization of our results and conclusions, we use NASA's EO-1 dataset from the Hyperion HS sensor. Our goal is to determine whether 1D-Justo-LiuNet continues to outperform 2D-CNNs, as observed in the HYPSO-1 dataset. As shown in Table~\ref{Table:generalisation_test_metrics_NASA_HYPERSPECTRAL}, we consider two 2D-CNNs. The first one is 2D-Justo-UNet-Simple, while the second one is nnU-Net 2D, a recent model that has shown promising results. The latter deep network~\cite{grabowski2022self} is a self-configuring nnU-Net, which was later compressed~\cite{grabowski2023squeezing} and presented in the competition On Cloud N: Cloud Cover Detection Challenge. After training, validating, and testing the 1D-CNN and 2D-CNN models on the EO-1 dataset, Table~\ref{Table:generalisation_test_metrics_NASA_HYPERSPECTRAL} confirms that 1D-Justo-LiuNet once again outperforms the 2D-CNNs. However, the table shows that this conclusion applies only when most channels are used. When the number of channels is drastically reduced, the spectral information loss is significant, and as expected, 1D-Justo-LiuNet does not outperform 2D models. However, the results indicate that the best segmentation performance is only achieved by 1D-Justo-LiuNet, with an accuracy of 0.82 and an F1 score of 0.97, particularly when a larger set of spectral channels is utilized. Therefore, we conclude that optimal segmentation in HS imagery is achieved when using 1D-CNNs, such as 1D-Justo-LiuNet, in conjunction with the largest possible spectral range. In short, our conclusions on HYPSO-1 are confirmed. A detailed description of our experiments on the NASA's EO-1 Hyperion dataset is provided in~\cite{kovac2024deep}.

Finally, Table~\ref{Table:generalisation_test_metrics_RGB_AERIAL} presents results from the RGB Satellite Imagery of Dubai (SID) dataset. The results on this 3-channel dataset indicate that our earlier conclusions generalize only to the hyperspectral domain. As expected, for satellite RGB imagery, 2D-CNNs and Fast-ViTs outperform 1D-CNNs, as each pixel contains only limited RGB bands, compared with the richer wavelength set found in HS data. In these cases, the lack of extensive spectral features needs to be compensated by incorporating neighboring spatial context.

\section{Conclusion} 
\label{Section: Conclusion}

In this work, we have trained and tested 20 models, including 1D-CNNs, 2D-CNNs and vision transformers, for segmentation of hyperspectral satellite imagery with a focus on lightweight models for future in-orbit edge inference. We have proposed a 1D-CNN, called 1D-Justo-LiuNet, which has exceeded expectations by outperforming all other state-of-the-art models. It has outclassed conventional 2D-CNN UNets, their lighter versions, as well as recent vision transformers for edge inference. 1D-Justo-LiuNet has achieved top accuracy with the smallest model size among all tested models. Our results have also been validated across multiple satellite datasets. First, we used the hyperspectral mission HYPSO-1 as the primary case study for pixel-level segmentation. We then conducted further generalization tests on NASA’s EO-1 hyperspectral mission, and obtained consistent results. We additionally evaluated the model on RGB Satellite Imagery of Dubai, where we confirmed our findings. We conclude that 1D-CNNs, specifically 1D-Justo-LiuNet and 1D-Justo-HuNet, are highly suitable for future in-orbit deployment. As future work, we propose deploying 1D-Justo-LiuNet in flight for optimal satellite operations.


\section*{Data availability}

The labeled dataset from HYPSO-1 used in this paper is openly available to download at \url{https://ntnu-smallsat-lab.github.io/hypso1_sea_land_clouds_dataset/}. For any issues with downloading the dataset, contact NTNU's HYPSO team. Additionally, our supplementary material is openly available at \url{https://github.com/NTNU-SmallSat-Lab/s_l_c_segm_hyp_img/}. Here, we provide the following resources. Firstly, the Python model codes for the 1D-CNNs, 2D-CNNs, and a reference to Apple's FastViT. The codes are accompanied by a notebook that illustrates how to perform inference on HYPSO-1 data, by 1D-CNN and 2D-CNN models. Secondly, there is an executable software program designed to serve as an oceanic-terrestrial-cloud segmenter for hyperspectral imagery acquired by the HYPSO-1 satellite. Thirdly, we provide trained models with learned parameters, with 46 distinct configurations, for the HYPSO-1 data. Fourth, we offer additional in-depth architectural details for the models. Fifth, we include information regarding the training, validation, and test data splits. Sixth, we share an extensive set of experimental figures, including multiple metrics. Seventh, we provide a deployment set consisting of 30 unlabeled HYPSO-1 images. Lastly, we present the generalization tests in the complete deployment set.


\section*{Acknowledgement}

We extend our gratitude to Anum Masood, Per Gunnar Kjeldsberg, and Simen Netteland.

\appendices  
\section{Direct Channel Reduction}
\label{Appendix:PCA_redution}
We obtain another HYPSO-1 dataset by reducing the 112 channels to merely 3 channels. To achieve this reduction, we explore the literature of dimensionality reduction extensively studied for HS satellite missions \cite{bakken2019effect, bakken2022development, penneindependent}. First, the focus on reducing dimensions originates from the need of having low-dimensional data to save computation resources, memory and power. Techniques such as Principal Component Analysis (PCA) \cite{rodarmel2002principal} and Independent Component Analysis (ICA) \cite{lupu2022stochastic} are proposed in the literature. In alignment with the state of the art, we employ PCA to reduce the 112 channels (normalized) to 3. To alleviate to a certain extent the computational demands of satellites operating with tight power constraints, we propose to conduct both the training and inference of PCA exclusively on ground. Eliminating PCA inference on the satellite enables potential power savings, which could be of importance to conduct the subsequent segmentation inference during flight. Therefore, and as also outlined by the literature \cite{giuffrida2021varphi}, we suggest implementing a direct channel selection (3 channels) on board, instead of deploying PCA inference. In this work, the purpose of employing PCA only on ground is to identify in advance the channels containing the most useful information. In particular, we use PCA to identify the channels that have the greatest influence on a single principal component, as determined by the weights of those channels. 
For the 112-channel HYPSO-1 dataset in L1b radiance, we find that the majority of influential channels are predominantly within the NIR range. However, given the necessity of considering other spectral ranges in addition to NIR for the adequate segmentation of sea, land, and clouds – as the NIR range alone is likely to be insufficient – we opt to pick the top three influential channels from the visible blue, the combined visible green and red, and finally, the NIR. In the dataset, the most influential channels within each spectral range are: 412.72 nm (channel 7) for the blue spectrum, 699.61 nm (channel 89) in the green and red spectra, and 747.77 nm (channel 103) in the NIR spectrum. These closely align with the bands often analyzed during HYPSO-1 operations, especially for channels 89 and 103.

\vspace{-3mm} 
\section{Data Dimensions Arrangement}
\label{Appendix:data_dimensions_arrengement}

Arranging the dimensions of the data is key before model training and inference, as models expect inputs in specific shapes. For 1D-CNN or 1D ML models, we transform each HS 3D data cube into a 2D array resembling tabular data. The models take inputs of the form BATCH SIZE $\times$ CHANNELS, where BATCH SIZE is the number of image pixels for which predictions are to be inferred. 1D-CNNs perform 1D convolutions separately on each pixel across its channels. For 2D-CNNs, optimizing memory and computational resources during training and inference is even more essential, and to help achieve this, we divide each data cube into smaller cubes, often referred as 3D patches. 
Padding extends the image borders to retain the integrity of edge information, and the padded borders can be eliminated in the final segmented image by cropping the extended spatial dimensions. After data padding, we proceed with non-overlapping patching. Unlike overlapping patching, this approach guarantees that each pixel is processed only once to reduce computational complexity. The input format for the 2D-CNNs is BATCH SIZE $\times$ PATCH SIZE $\times$ PATCH SIZE $\times$ CHANNELS, where BATCH SIZE is the number of 3D patches. We find that a PATCH SIZE of 48 is suitable for our experiments as it aligns with the encoder-decoder networks, which repeatedly decrease the spatial dimensions by a factor of 2 via pooling. Therefore, selecting a small PATCH SIZE of 48 ensures that the spatial dimensions subjected to pooling remain integer as the network's depth increases. This is also convenient for reducing computational complexity in orbit, as opposed to processing larger patches.

\bibliography{main.bib} 

\vspace{-15mm} 
\begin{IEEEbiography}[{\includegraphics[width=1in,height=1.25in,clip,keepaspectratio]{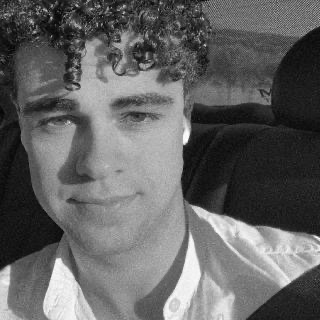}}]{Jon Alvarez Justo} graduated with honors in 2018 as a Telecommunications Technologies Engineer at the University of Vigo, Spain. He has expertise in various disciplines within Information and Communication Technologies, including computer science (software and hardware development, data science and analytics, signal processing), telematics, and telecommunications. After completing his Master's degree in 2020 at the Norwegian University of Science and Technology (NTNU), he started pursuing a doctoral degree. His research focuses on studying, developing, and verifying AI algorithms for deployment in space satellites to increase platform automation and optimize operations. His work is part of the NTNU Small Satellites Laboratory, which leads the HYPSO space missions for Earth observation.
\end{IEEEbiography}

\vspace{-15mm} 
\begin{IEEEbiography}[{\includegraphics[width=1in,height=1.25in,clip,keepaspectratio]{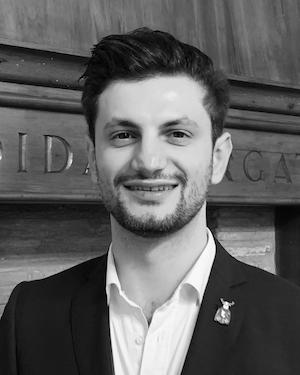}}]{Alexandru Ghi\c{t}\u{a}} received the B.Sc. and M.Sc. in Mathematics from the University of Bucharest in 2016 and 2018, respectively. He is pursuing a PhD at the Faculty of Mathematics and Computer Science at the University of Bucharest, with the main research area of developing efficient learning loss functions for deep learning models.
\end{IEEEbiography}

\vspace{-15mm} 
\begin{IEEEbiography}[{\includegraphics[width=1in,height=1.25in,clip,keepaspectratio]{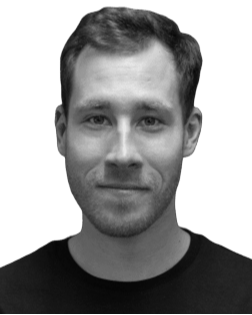}}]{Daniel~Ková\v{c}} is a Data Scientist at Scicake, a researcher at the Brain Diseases Analysis Laboratory (BDALab), and a PhD student at Brno University of Technology, where he specializes in digital signal processing, statistics, and machine learning. He holds a Master's degree in Audio Engineering from the same institution and has completed research internships at the University of Vienna and Afeka Tel Aviv Academic College of Engineering. His research focuses on developing digital biomarkers for the objective assessment of speech and language disorders in neurodegenerative diseases. He also explores image segmentation and classification methods using hyperspectral satellite data, with applications in precision agriculture and environmental monitoring.

\end{IEEEbiography}

\vspace{-15mm} 
\begin{IEEEbiography}[{\includegraphics[width=1in,height=1.25in,clip,keepaspectratio]{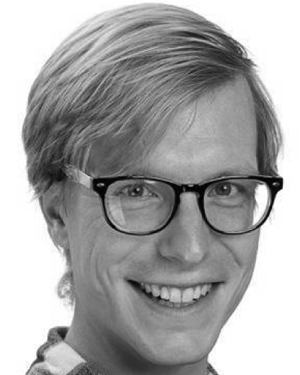}}]{Joseph L. Garrett} (Member, IEEE) received the B.Sc. degree in physics and mathematics from The Ohio State University, Columbus, OH, USA, in 2011, and the Ph.D. degree in physics from the University of Maryland, College Park, MD, USA, in 2017. His work is on hyperspectral imaging and image processing from satellites and drones as a Postdoctoral Researcher at the Norwegian University of Science and Technology (NTNU), Trondheim, Norway.
\end{IEEEbiography}

\vspace{-8mm} 
\begin{IEEEbiography}[{\includegraphics[width=1in,height=1.25in,clip,keepaspectratio]{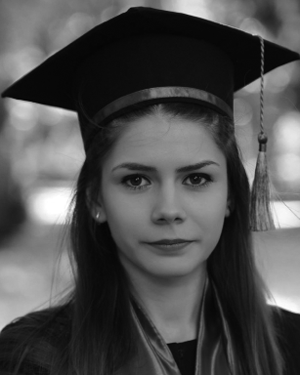}}]{Mariana-Iuliana Georgescu} received the BSc degree from the Faculty of Mathematics and Computer Science and the MSc degree in artificial intelligence from University of Bucharest, in 2017 and 2019, respectively. She has recently received in 2024 her PhD degree at the Faculty of Mathematics and Computer Science, University of Bucharest. Although she is early in her research career, she is the first author of six papers published at conferences and journals. Her research interests include artificial intelligence, computer vision, machine learning, deep learning, and medical image processing.
\end{IEEEbiography}

\vspace{-8mm} 
\begin{IEEEbiography}[{\includegraphics[width=1in,height=1.25in,clip,keepaspectratio]{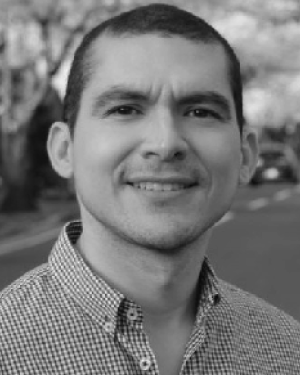}}]{Jesus Gonzalez-Llorente} (Senior Member, IEEE) received his B.Sc. in electronics engineering from the National Univ. of Colombia, 2003, the DESS. in software engineering from the District Univ. of Bogota, 2005, the M.Sc. degree in electrical engineering from the Univ. of Puerto Rico at Mayagüez, 2009, and the Ph.D. in engineering from the Kyushu Institute of Technology, Japan, in 2019. He was a researcher with the Arecibo Observatory under Cornell Univ. operation, 2010. He was also a visiting scholar at the Univ. of Arkansas, USA.  He was an IEEE AESS Distinguished Lecturer (2023–2024). He is a visiting researcher at the Nihon Univ., Japan. His research interests include solar photovoltaic systems, small satellites, and space systems engineering. Dr. Gonzalez-Llorente received the Emerging Space Leader award from the International Astronautical Federation (IAF), in 2015. He also received the United Nations/Japan Long-Term Fellowship, from 2016 to 2019, funded by UNOOSA and MEXT, Japan. He has been recognized as an Associate Researcher by the Ministry of Science Technology and Innovation of Colombia, since 2022.
\end{IEEEbiography}

\vspace{-24mm} 
\begin{IEEEbiography}[{\includegraphics[width=1in,height=1.25in,clip,keepaspectratio]{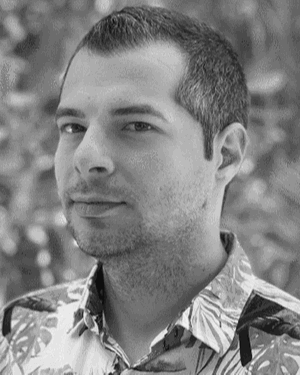}}]{Radu Tudor Ionescu} (Member, IEEE) received the PhD degree from the University of Bucharest, in 2013. He is a professor with the University of Bucharest, Romania. He received the 2014 Award for Outstanding Doctoral Research from the Romanian Ad Astra Association. His research interests include machine learning, computer vision, image processing, computational linguistics, and medical imaging. Dr. Ionescu published more than 130 articles at international venues (including CVPR, NeurIPS, ICCV, ACL, SIGIR, EMNLP, NAACL, TPAMI, IJCV, CVIU), and a research monograph with Springer. He received the ``Caianiello Best Young Paper Award'' at ICIAP 2013.
\end{IEEEbiography}

\vspace{-24mm} 
\begin{IEEEbiography}[{\includegraphics[width=1in,height=1.25in,clip,keepaspectratio]{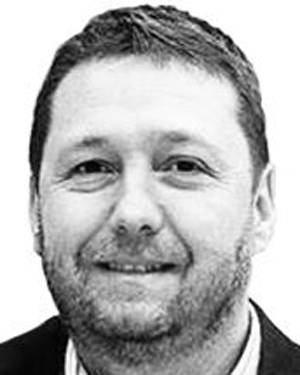}}]{Tor Arne Johansen} (Senior Member, IEEE) received the M.Sc. and Ph.D. degrees in electrical and computer engineering from the Norwegian University of Science and Technology (NTNU), Trondheim, Norway, in 1989 and 1994, respectively. From 1995 to 1997, he worked with SINTEF, Trodheim, Norway, as a Researcher before he was appointed Associated Professor with the Norwegian University of Science and Technology, in 1997, and Professor in 2001. He has authored or coauthored more than hundred articles in the areas of control, estimation, and optimization with applications in the marine, aerospace, automotive, biomedical, and process industries. In 2002, he co-founded the company Marine Cybernetics AS, Trodheim, Norway, where he was Vice President until 2008. He is the Director of the Unmanned Aerial Vehicle Laboratory, NTNU, and the SmallSat Laboratory, NTNU. He recently co-founded the spin-off companies Scout Drone Inspection, UBIQ Aerospace, Zeabuz, and SentiSystems. He is currently a Principal Researcher with the Center of Excellence on Autonomous Marine Operations and Systems, NTNU (NTNU-AMOS). Dr. Johansen was the recipient of the 2006 Arch T. Colwell Merit Award of the SAE.
\end{IEEEbiography}

\end{document}